\documentclass{article}

\usepackage[preprint, nonatbib]{neurips_2024}

\usepackage{amssymb} 
\usepackage{graphicx}
\usepackage{mdframed}
\usepackage{subcaption}
\usepackage{algpseudocode}
\usepackage{algorithm}
\usepackage{bm}
\usepackage{amsmath}

\usepackage[utf8]{inputenc} 
\usepackage[T1]{fontenc}    
\usepackage{hyperref}       
\usepackage{url}            
\usepackage{booktabs}       
\usepackage{amsfonts}       
\usepackage{nicefrac}       
\usepackage{microtype}      
\usepackage{xcolor}         

\usepackage[many]{tcolorbox}    	
\definecolor{main}{HTML}{070000}    
\definecolor{sub}{HTML}{eeeeee}     

\newtcolorbox{boxH}{
    colback = sub, 
    colframe = main, 
    boxrule = 0pt, 
    leftrule = 6pt 
}

\title{How Do Transformers ``Do'' Physics? Investigating the Simple Harmonic Oscillator}

\author{%
  Subhash Kantamneni, Ziming Liu$^\dagger$, \& Max Tegmark$^\dagger$\\
  Massachusetts Institute of Technology\\
  \texttt{\{subhashk,zmliu,tegmark\}@mit.edu} \\
}

\begin{document}

\maketitle\def\thefootnote{$\dagger$}\footnotetext{equal advising.}\def\thefootnote{\arabic{footnote}}

\begin{abstract}
   How do transformers model physics? Do transformers model systems with interpretable analytical solutions, or do they create ``alien physics'' that is difficult for humans to decipher? We take a step in demystifying this larger puzzle by investigating the simple harmonic oscillator (SHO), $\ddot{x}+2\gamma \dot{x}+\omega_0^2x=0$, one of the most fundamental systems in physics. Our goal is to identify the methods transformers use to model the SHO, and to do so we hypothesize and evaluate possible methods by analyzing the encoding of these methods' intermediates. We develop four criteria for the use of a method within the simple testbed of linear regression, where our method is $y = wx$ and our intermediate is $w$: (1) Can the intermediate be predicted from hidden states?
   (2) Is the intermediate's encoding quality correlated with  model performance? 
   (3) Can the majority of variance in hidden states be explained by the intermediate?
   (4) Can we intervene on hidden states to produce predictable outcomes? 
   Armed with these two correlational (1,2), weak causal (3) and strong causal (4) criteria, we determine that transformers use known numerical methods to model trajectories of the simple harmonic oscillator, specifically the matrix exponential method. Our analysis framework can conveniently extend to high-dimensional linear systems and nonlinear systems, which we hope will help reveal the ``world model'' hidden in transformers.
   
\end{abstract}

\section{Introduction}


Transformers are state of the art models on a range of tasks \cite{vaswani2017allyouneed, Devlin2019BERTPO, chowdhery2022palm, liu2021cvtransformer}, but our understanding of how these models represent the world is limited. Recent work in mechanistic interpretability \cite{elhage2022superposition, olsson2022context, liu2022towards,elhage2021mathematical, chughtai2023toy, gurnee2023finding, wang2023interpretability, conmy2023towards} has shed light on how transformers represent mathematical tasks like modular addition \cite{nanda2023progress,zhog2023clock,liu2022towards}, yet little work has been done to understand how transformers model physics. This question is crucial, as for transformers to build any sort of ``world model,'' they must have a grasp of the physical laws that govern the world \cite{liu2024physcontext}\footnote{Transformers with better ``world models'' are also at greater risk of misuse.}.

Our key research question is:  How do transformers model physics? This question is intimidating, since even humans have many different ways of modeling the same underlying physics \cite{Shapiro2010}. In the spirit of hypothesis testing, we reformulate the question as: given a known modeling method $g$, does the transformer learn $g$? If a transformer leverages $g$, its hidden states must encode information about important intermediate quantities in $g$. We focus our study on the simple harmonic oscillator $\ddot{x}+2\gamma \dot{x} + \omega_0^2 x=0$, where $\gamma$ and $\omega_0$ are the damping and frequency of the system respectively. Given the trajectory points $\{(x_0,v_0), (x_1,v_1), \dots, (x_n,v_n)\}$ at discrete times $\{t_0, t_1, \dots, t_n\}$, we task a transformer with predicting $(x_{n+1}, v_{n+1})$ at time $t_{n+1}$, as shown in Fig. \ref{fig:Figure1}. In this setting, $g$ could be a numerical simulation the transformer runs after inferring $\gamma, \omega_0$ from past data points. We would then expect some form of $\gamma$ and $\omega_0$ to be intermediates encoded in the transformer. How can we show that intermediates and the method $g$ are being used? 

We develop criteria to demonstrate the transformer is using $g$ by studying intermediates in a simpler setting: in-context linear regression, $y = wx$. As correlational evidence for the model's internal use of $w$, we find that the intermediate $w$ can be encoded linearly, nonlinearly, or not at all. We also link the performance of models to their encoding of $w$ and use it as an explanation for in-context learning. We generate causal evidence for the use of $w$ by analyzing how much of the hidden states' variance $w$ explains and linearly intervening on the network to predictably change its behavior. 

We use these developed criteria of intermediates to study how transformers model the simple harmonic oscillator (SHO), a fundamental model in physics. We generate multiple hypotheses for the method(s) transformers use to model the trajectories of SHOs, and use our criteria from linear regression to show correlational and causal evidence that transformers employ known numerical methods, specifically the matrix exponential, to model trajectories of SHOs. Although our analysis is constrained to the SHO in this paper, our framework naturally extends to some high-dimensional linear and nonlinear systems.

\textbf{Organization} In Section \ref{sec:linreg} we define and investigate intermediates in the setting of linear regression and use this to develop criteria for transformers' use of a method $g$. In Section \ref{sec:sho} we hypothesize that transformers use numerical methods to model the SHO, and use our criteria of intermediates to provide causal and correlational evidence for transformers' use of the matrix exponential.  Due to limited space, related work is deferred to Appendix \ref{sec:relatedwork}.

\section{Developing Criteria for Intermediates with Linear Regression}
\label{sec:linreg}

Our main goal is to determine which methods transformers use to model the simple harmonic oscillator. We would like to do this by generating criteria based on the encoding of relevant intermediates. In this section, we develop our criteria of intermediates in a simpler setting: linear regression. Notably, linear regression is identical to predicting the acceleration from the position of an undamped harmonic oscillator ($\gamma = 0$), making this setup physically relevant.

\begin{figure*}
  \includegraphics[width = 1\textwidth]{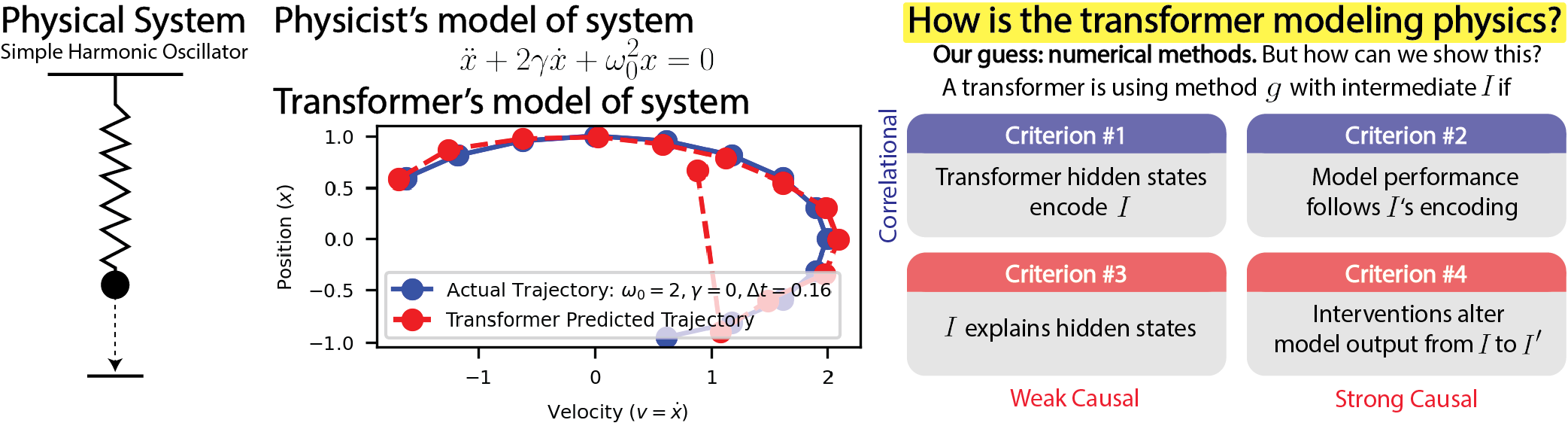}
  \caption{We aim to understand how transformers model physics through the study of meaningful intermediates. We train transformers to model simple harmonic oscillator (SHO) trajectories and use our developed criteria of intermediates to show that transformers use known numerical methods to model the SHO.}
  \label{fig:Figure1}
\end{figure*}

\textbf{Setup} In our linear regression setup, we generate $\bm{X}$ and $\bm{w}$ between $[-0.75,0.75]$, where $\bm{X}$ has size $(5000,65)$ and $\bm{w}$ has size $(5000,)$. We generate $\bm{Y}=\bm{wX}$, and train transformers to predict $y_{n+1}$ given $\{x_1,y_1,...,x_n,y_n,x_{n+1}\}$.

Since in-context linear regression is well studied for transformers \cite{Akyrek2022WhatLA, garg2022icl}, we use this simple setting to ask and answer fundamental questions about intermediates, namely:
\begin{itemize}
    \item \textbf{What} is an intermediate?
    \item \textbf{How} can intermediates be encoded and how can we robustly probe for them?
    \item \textbf{When}, or under what circumstances, are intermediates encoded?
\end{itemize}

All of these questions develop an understanding of intermediates that builds up to the \textbf{Key Question: How can we use intermediates to demonstrate that a transformer is actually using a method in its computations?} By answering this question for linear regression, we generate four correlational and causal criteria to demonstrate a transformer is using a method in its computations, which we can then apply to understand the simple harmonic oscillator, as shown in Fig. \ref{fig:Figure1}.

\subsection{What is an intermediate?}

We define an intermediate as a quantity that a transformer uses during computation, but is not a direct input/output to/of the transformer. More formally, if the input to the transformer is $\bm{X}$ and its output is $\bm{Y}$, we can model the transformer's computation as $\bm{Y} = g(\bm{X}, I)$, where $g$ is the method used and $I$ is the intermediate of that method. For example, if we want to determine if the transformer is computing the linear regression task using $\bm{Y} = \bm{wX}$, then $I=\bm{w}$, $g(\bm{X},I) = g(\bm{X}, \bm{w}) = \bm{wX}$.

\subsection{How can intermediates be encoded and how can we robustly probe for them?}

We want to understand what form of the intermediate, $f(I)$, is encoded in the network's hidden states. For example, while it may be obvious to humans to compute $y=wx$, perhaps transformers prefer $\exp(\log(w) + \log(x))$ or $\sqrt{w^2x^2}$. We want to develop a robust probing methodology that captures these diverse possibilities. We identify three ways an intermediate $I$ can be represented: linearly encoded, nonlinearly encoded, and not encoded at all. We use $HS$ to mean hidden state.

\textbf{Linearly encoded} We say $I$ is linearly encoded in a hidden state $HS$ if there is a linear network that takes $I = \text{Linear}(HS)$. We determine the strength of the linear encoding by evaluating how much of the variance in $I$ can be explained by $HS$, i.e. the $R^2$ of the probe.

\textbf{Nonlinearly encoded} To probe for an arbitrary $f(I)$, we define a novel \textbf{Taylor probe}, which finds coefficients $a_i$ such that $f(I) = a_1I + a_2I^2+...+a_nI^n$, and $f(I) =  \text{Linear}(HS)$. 
To actually implement this probing style, we use Canonical Correlation Analysis probes, which given some multivariate data $X$ and $Y$, finds directions within $X$ and $Y$ that are maximally correlated \cite{cca}. Here, $X= [I, I^2, I^3,...,I^n]$, and $Y=HS$. If $I$ is of bounded magnitude and $n$ is sufficiently large, we are able to probe the transformer for any function $f(I)$. In practice, we use $n \leq 5$.

\textbf{Not encoded}
If $I$ fails to be linearly or nonlinearly encoded, we say that it is not encoded within the network. For example, there are at least two ways to predict $y_2$ from $\{x_1,y_1,x_2\}$ such that $y_2=\frac{y_1}{x_1}x_2$. (1) $w=y_1/x_1$ is encoded, and $y_2=wx_2$. (2) $w'=x_2/x_1$ is encoded (so $w=y_1/x_1$ is not encoded), and $y_2 = w'y_1$. Thus, it is not guaranteed that $w$ is encoded.


\subsection{When, or under what circumstances, are intermediates encoded?}

We want to apply our probing techniques to better understand what type of models generate intermediates. Under the described setting of linear regression, we train transformers of size $L = [1,2,3,4,5]$ and $H = [2,4,8,16,32]$ \footnote{All transformers trained in this study use one attention head and no LayerNorm to aid interpretability, and are trained on a NVIDIA Volta GPU with the hyperparameters $\rm{epochs} = 20000, \rm{lr} = 10^{-3}, \rm{batch size} = 64$ using the Adam optimizer \cite{Kingma2014AdamAM}.}, where $L$ is the number of layers and $H$ is the hidden size of the transformer. We find that these models generalize to out-of-distribution test data ($0.75\leq|\bm{w}|\leq1$) in Appendix Fig. \ref{fig:linregtest}, but we focus on investigating intermediates on in-distribution training data.

\begin{figure}
    \centering
    \includegraphics[width = \textwidth]{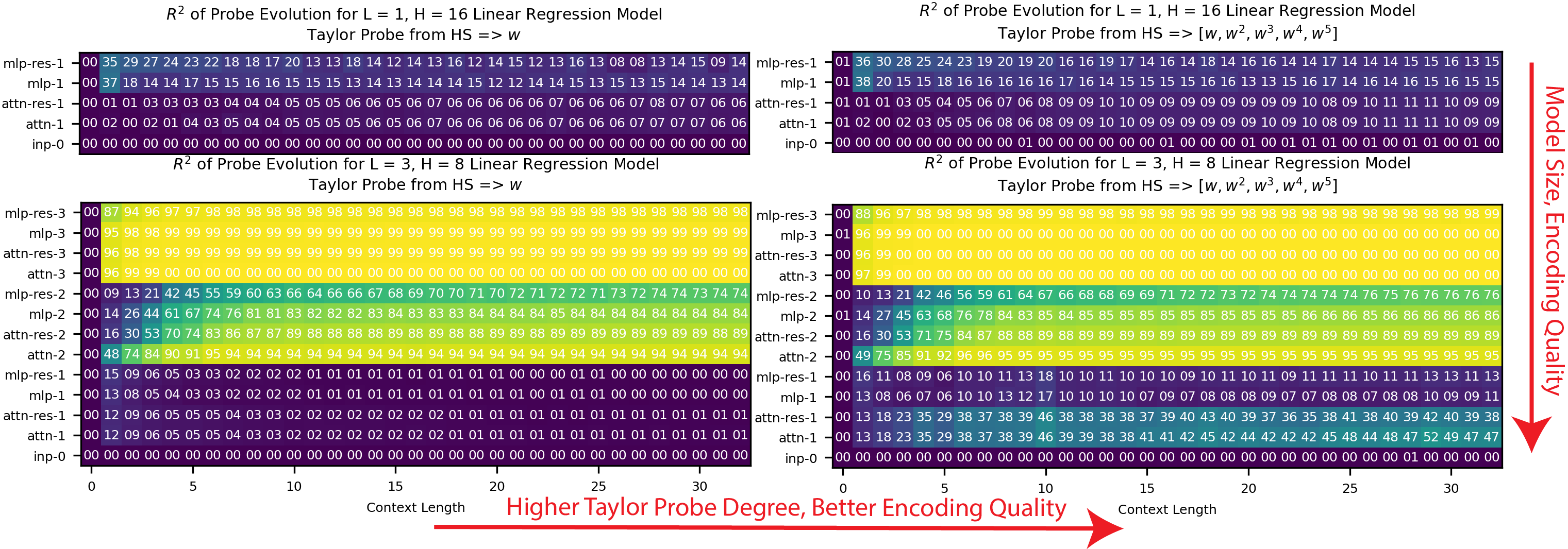}
    \caption{We plot the $R^2$ of Taylor probes for the intermediate $\bm{w}$ within models trained on the task $\bm{Y} = \bm{wX}$. We see that larger models have $\bm{w}$ encoded, often linearly, with little gain as we move to higher degree Taylor probes, while smaller models do not have $\bm{w}$ encoded.}
    \label{fig:probeevolve}
\end{figure}

\textbf{Larger models have stronger encodings of intermediates} We find that smaller models often don't have $\bm{w}$ encoded, while larger models encode $\bm{w}$ linearly, as evidenced by Fig. \ref{fig:probeevolve}. We formalize this further by defining $\max(\bar{R^2})$ as the maximum value taken over depth positions of the mean $R^2$ of $\bm{w}$ probes taken over context length. In Appendix Fig. \ref{fig:phasetransition}, we observe a clear phase transition in encoding across model size, and also find that $\max(\bar{R^2})$ does not significantly improve if we extend the degree of the Taylor probes to $n>2$. Thus, in the case of linear regression, we find that models represent $\bm{w}$ linearly, quadratically, or not at all.

We attribute the stronger encoding of $\bm{w}$ in larger models to the "lottery ticket hypothesis" - larger models have more "lottery tickets" in their increased capacity to find a "winning" representation of $\bm{w}$ \cite{Frankle2018TheLT, Liu2023ANS}. Interestingly, the intuitive understanding that larger models have $\bm{w}$ better encoded leads us to the counterintuitive conclusion that larger models are actually \textit{more} interpretable for our purposes.

\begin{figure*}
  \includegraphics[width = 1\textwidth]{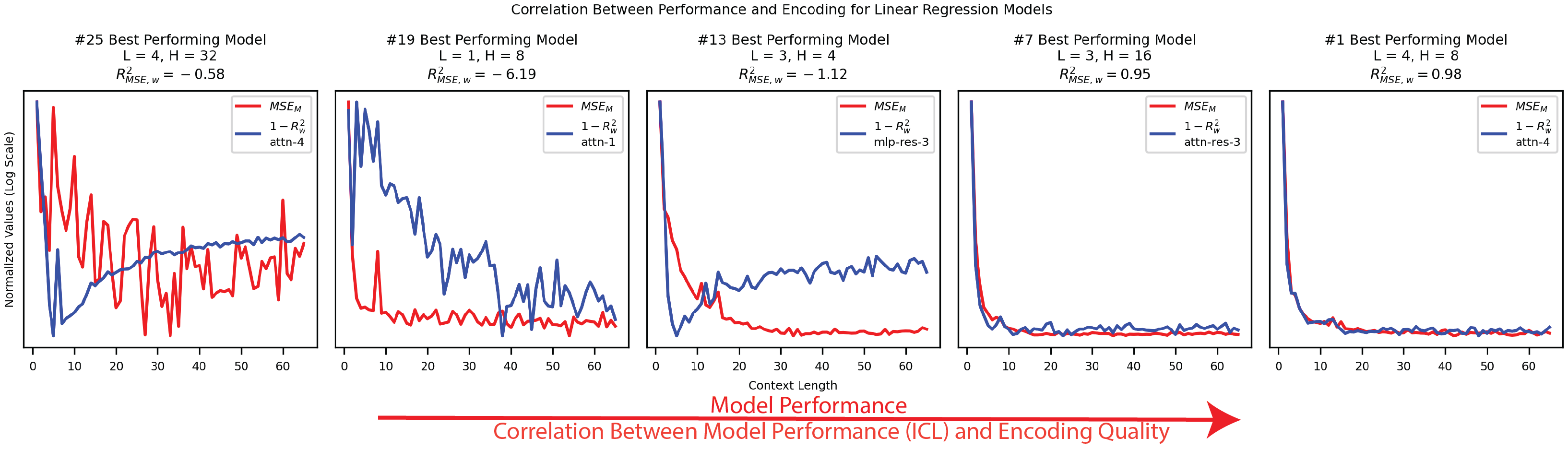}
  \caption{We test the correlation between model performance and the encoding of $\bm{w}$ on 5 of our 25 models of evenly spaced performance quality. We plot normalized values for the error of the encoding ($1-R_w^2$) in red and the mean squared error of the model (${MSE}_M$) in blue. We find that the ability of the best performing models to in-context learn is highly correlated with their encoding of $\bm{w}$ ($R^2(MSE,w)$.}
  \label{fig:ICL_wenc}
\end{figure*}

\textbf{Encoding quality is tied to model performance} In Appendix Fig. \ref{fig:linregmser2corr}, we find that better performing models generally have stronger encodings of $\bm{w}$. In Fig. \ref{fig:ICL_wenc}, we also find that the improvements in model prediction as a function of context length, or in-context learning, are correlated to improvements in $\bm{w}$'s encoding, which we would expect if our models were using $\bm{w}$ in their computations.

\subsection{Key Question: How can we use intermediates to demonstrate that a transformer is actually using a method in its computations?}

So far we have discovered that models encode $\bm{w}$, either linearly or nonlinearly, and found relationships between model size, performance, and encoding strength. But how can we ensure that the model is actually using $\bm{w}$ in its computations and the encoding of $\bm{w}$ is not just a meaningless byproduct \cite{ravichander2020badprobes}?

\begin{figure*}[ht!]
\centering
  \includegraphics[width=0.25\textwidth]{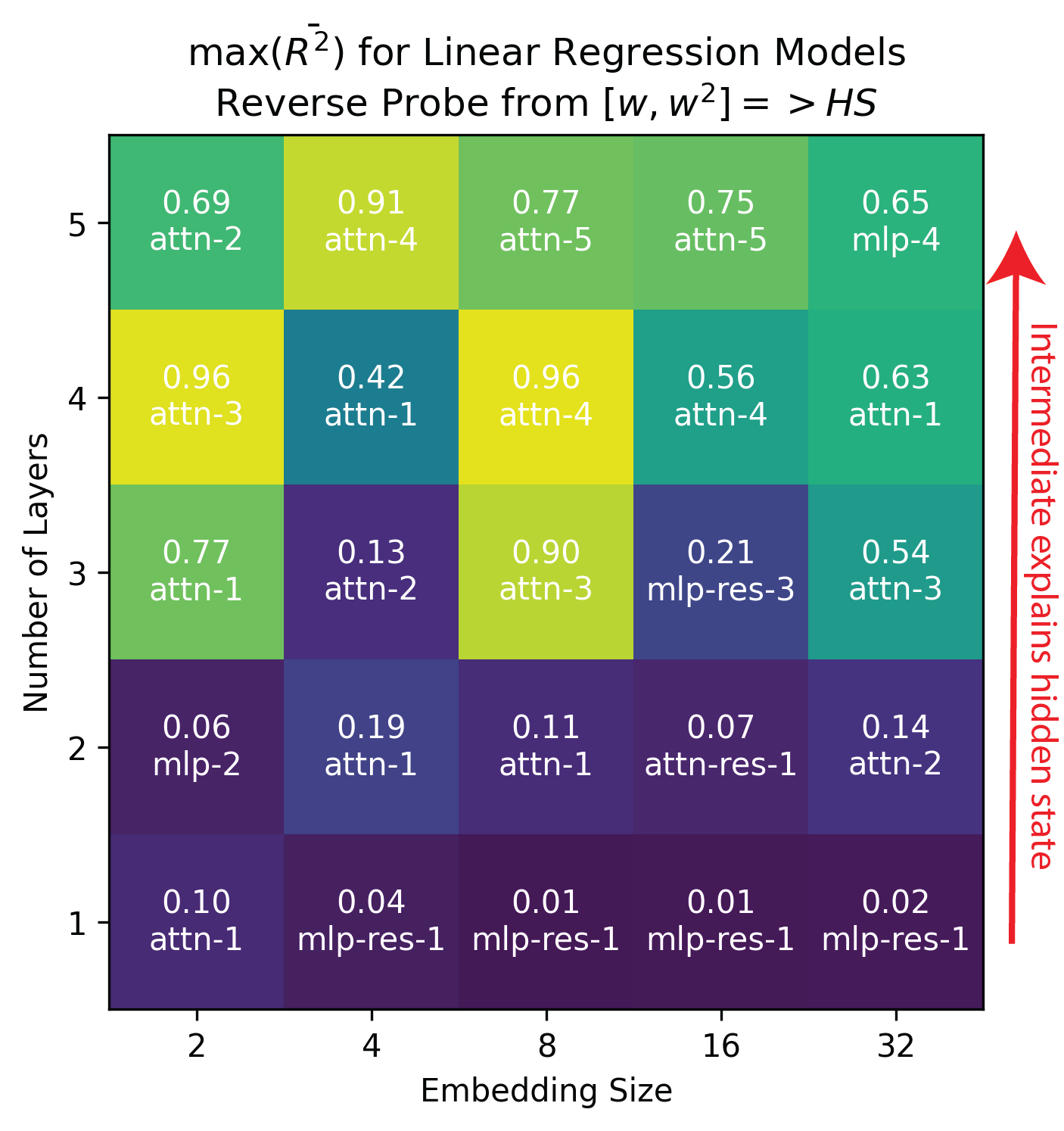}
  \includegraphics[width=0.7\textwidth]{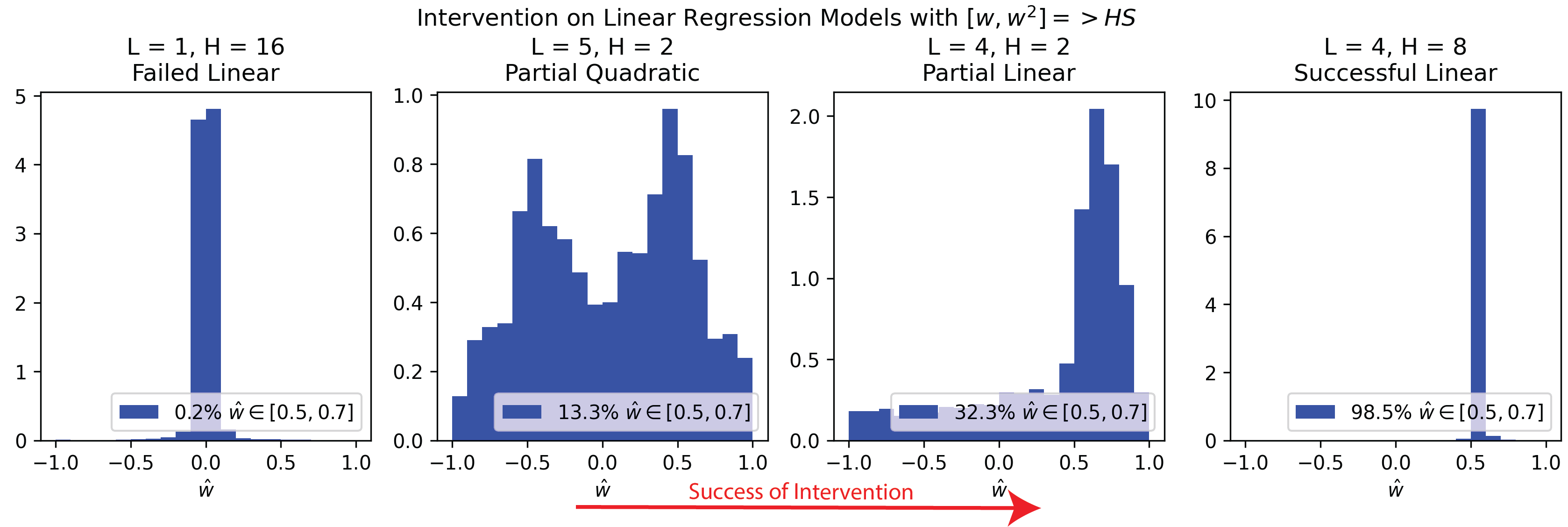}

  \caption{Left: We plot $\max({\bar{R^2}})$ of the reverse probe from $[w, w^2] \rightarrow  HS$ across all models, and find that the intermediate $\bm{w}$ can explain significant amounts of variance in model hidden states. \label{fig:reverseprobe}
    \newline
    Right: We intervene using reverse probes to make all models output $\bm{w'} = 0.5$. This intervention can either fail (16/25), be partially successful nonlinearly (2/25) or linearly (3/25), or be successful (4/25). \label{fig:wintervention}
  }
\end{figure*}

\textbf{Reverse Probing} To ensure that $\bm{w}$ is not encoded in some insignificant part of the residual stream, we set up probes going from $[w,w^2]\rightarrow HS$, as opposed to $HS\rightarrow f(w)$. In Fig. \ref{fig:reverseprobe}, we often find that $\bm{w}$ can explain large amounts of variance in model hidden states, implying that these hidden states are dedicated to representing $\bm{w}$. We take this as weak causal evidence that $\bm{w}$ is being used by the model - otherwise, it is unclear why a part of the model would be dedicated to storing $\bm{w}$.



\textbf{Intervening} We can also use reverse probes to intervene on the models' hidden states and predictably change their output from $\bm{w}\rightarrow \bm{w'}$. In Fig. \ref{fig:wintervention} we attempt to make $\bm{w'} = 0.5$ for all series, and then measure the observed $\bm{\hat{w}}$ from the models' outputs ($\hat{w} = \hat{y}_n/x_n$). For 4 out of 25 models the intervention worked, providing strong causal evidence that the model uses its internal representation of $\bm{w}$ in computations. For models where we identified a quadratic representation of $\bm{w}$, we see that $\bm{w} = 0.5, -0.5$ are both represented in the observed intervention.

\textbf{Putting it all together} We can generalize our understanding of intermediates from linear regression to create criteria for a transformer's use of a method $g$ in its computations.

\begin{boxH}
\textbf{Criteria for use of a method $g$ with an associated, unique intermediate $I$}
\begin{enumerate}
    \item If a model uses a method $g$, its hidden states should encode $I$ (shown in Fig. \ref{fig:probeevolve}).
    \item If a model uses a method $g$, model performance should improve if $I$ is better represented (shown in Fig. \ref{fig:ICL_wenc}).
    \item If and only if the model uses $g$, we expect some hidden state's variance to be almost fully explained by $I$ (shown in Fig. \ref{fig:reverseprobe}). 
    \item If and only if the model uses $g$, we can intervene on hidden states to change $I\rightarrow I'$ and predictably change the model output from $g(\bm{X}, I) \rightarrow  g(\bm{X},I')$ (shown in Fig. \ref{fig:wintervention})
\end{enumerate}

\end{boxH}

The first two criteria for a transformer's use of $g$ are correlational and the last two are weak and strong causal. Using these criteria (summarized in Fig. \ref{fig:Figure1}), we can now investigate how transformers model more complex systems like the simple harmonic oscillator.

\section{Investigating the Simple Harmonic Oscillator}
\label{sec:sho}

We now apply our developed criteria of intermediates to investigate how transformers represent physics, specifically the methods they use to model the simple harmonic oscillator (SHO). The simple harmonic oscillator is ubiquitous in physics, used to describe phenomena as diverse as the swing of a pendulum, molecular vibrations, the behavior of AC circuits, and quantum states of trapped particles. Given a series of position and velocity data of a simple harmonic oscillator at a sequence of timesteps, we ask 
\begin{enumerate}
    \item Can a transformer successfully predict the position/velocity at the SHO's next timestep?
    \item Can we determine what computational method the transformer is using in this prediction?
\end{enumerate}

\subsection{Mathematical and computational setup}
The simple harmonic oscillator is governed by the linear ordinary differential equation (ODE)
\begin{equation}
    \ddot{x} + 2\gamma \dot{x} + \omega_0^2x = 0.
    \label{eq:DHO}
\end{equation}
The two physical parameters of this equation are $\gamma$, the damping coefficient, and $\omega_0$, the natural frequency of the system. An intuitive picture for the SHO is a mass on a spring that is pulled from its equilibrium position by some amount $x_0$ and let go, as visualized in Fig. \ref{fig:Figure1}. $\omega_0$ is related to how fast the system oscillates, and $\gamma$ is related to how soon the system decays to equilibrium from the internal resistance of the spring. We focus on studying how a transformer models the undamped harmonic oscillator, where $\gamma = 0$. Given some initial starting position ($x_0$), velocity ($v_0$), and timestep $\Delta t$, the time evolution of the undamped harmonic oscillator is
\begin{equation}
    \begin{aligned}
        x_k &= x_0 \cos (k \omega_0 \Delta t) +  \frac{v_0}{\omega_0} \sin (k \omega_0 \Delta t) \\
        v_k &= v_0 \cos (k \omega_0 \Delta t) - \omega_0 x_0 \sin (k \omega_0 \Delta t),
    \end{aligned}
    \label{eq:SHO_solutions}
\end{equation}
where $v = \frac{dx}{dt}$. We generate 5000 timeseries of 65 timesteps for various values of $\omega_0, \Delta t, x_0, $ and $v_0$, described in Appendix \ref{sec:uho_app}. Following the procedure for linear regression, we train transformers of size $L = [1,2,3,4,5]$ and $H = [2,4,8,16,32]$ to predict $(x_{n+1},v_{n+1})$ given $\{(x_0,v_0),(x_1,v_1),...(x_n,v_n)\}$. In Appendix Fig. \ref{fig:undampedSHOdata}, we see that our transformers are able to accurately predict the next timestep in the timeseries of out-of-distribution test data, and this prediction gets more accurate with context length (i.e. in-context learning). But how is the transformer modeling the simple harmonic oscillator internally?


\subsection{What methods could the transformer use to model the simple harmonic oscillator?}

Human physicists would model the simple harmonic oscillator with the analytical solution to Eq. \ref{eq:DHO}, but it is unlikely that a transformer would do so. Transformers are numerical approximators which use statistical patterns in data to make predictions, and in that spirit, we hypothesize transformers use numerical methods to model SHOs. There is a rich literature on numerical methods that approximate solutions to linear ordinary differential equations \cite{rkmethod, lmmethod, uvic2024matexp}, and we highlight three possible methods the transformer could be using in our theory hub. For notation, we note that Eq. \ref{eq:DHO} can be written as 
\begin{equation}
    \begin{bmatrix}
        \dot{x} \\
        \dot{v}
    \end{bmatrix} = 
    \begin{bmatrix}
        0 & 1 \\
        -\omega_0^2 & -2 \gamma
    \end{bmatrix}
    \begin{bmatrix}
        x \\
        v
    \end{bmatrix} = 
\bm{A} \begin{bmatrix}
    x \\
    v
\end{bmatrix}.
\end{equation}

\begin{table}[]
\centering
\begin{tabular}{c|l|c}
\textbf{Method}            & \textbf{$\bm{g(X,I)}$} & $\bm{I}$ \\ \hline
Linear Multistep           & $\displaystyle \begin{bmatrix}
    x_{n+1} \\
    v_{n+1}
\end{bmatrix} = \sum_{j=0}^k \alpha_j \begin{bmatrix} x_{n-j} \\ v_{n-j} \end{bmatrix} +  \sum_{j=-1}^k \beta_j \bm{A} \Delta t \begin{bmatrix} x_{n-j} \\ v_{n-j} \end{bmatrix}$ & $\bm{A} \Delta t$ \\ \hline
Taylor Expansion           & $\displaystyle \begin{bmatrix}
    x_{n+1} \\
    v_{n+1}
\end{bmatrix} = \sum_{j=0}^{k} \bm{A}^j \frac{\Delta t^j}{j!} 
    \begin{bmatrix}
    x_n \\
    v_n
    \end{bmatrix}$                                                                           & $(\bm{A} \Delta t)^j$ \\ \hline
Matrix Exponential         & $\displaystyle \begin{bmatrix}
    x_{n+1} \\
    v_{n+1}
\end{bmatrix} = e^{\bm{A} \Delta t} \begin{bmatrix}
    x_n \\
    v_n
\end{bmatrix}$                                                                                     & $e^{\bm{A} \Delta t}$ \\
\end{tabular}
\caption{Our theory hub of numerical methods and relevant intermediates transformers could be using to model the simple harmonic oscillator.}
\label{tab:num_methods}
\end{table}

\textbf{Linear Multistep Method}
Our model could be using a linear multistep method, which uses values of derivatives from several previous timesteps to estimate the future timestep.  We describe the $k$th order linear multistep method in Table \ref{tab:num_methods} with coefficients $\alpha_j$ and $\beta_j$.

\textbf{Taylor Expansion Method}
The model could also be using higher order derivatives from the previous timestep to predict the next timestep \footnote{This is equivalent to the nonlinear single step Runge-Kutta method for a homogenous linear ODE with constant coefficients}. We describe the $k$th order Taylor expansion in Table \ref{tab:num_methods}.

\textbf{Matrix Exponential Method} While the two methods presented above are useful approximations for small $\Delta t$, the matrix exponential uses a $2 \times 2$ matrix to exactly transform the previous timestep to the next timestep. We describe it in Table \ref{tab:num_methods}. This method is the $\lim_{k \to \infty}$ of the Taylor expansion method.

In order to use the criteria described in Section \ref{sec:linreg} to figure out which method(s) our model is using, we need to define relevant intermediates for each method $g$. Similarly to linear regression, the intermediates are the coefficients of the input, but are now $2 \times 2$ matrices and not a single value. We summarize our methods and intermediates in our theory hub in Table \ref{tab:num_methods}. Notably, these methods are viable for any homogeneous linear ordinary differential equation with constant coefficients, and potentially non linear differential equations as well (see Appendix \ref{sec:generalizingtoothersystems}).

\subsection{Evaluating methods for the undamped harmonic oscillator}
We apply the four criteria established for linear regression (Fig. \ref{fig:Figure1}) to evaluate if transformers use the methods in Table \ref{tab:num_methods}. For the Taylor expansion intermediate, we use $j = 3$ to distinguish it from the linear multistep method, although our results are generally robust to $j\leq5$ (Appendix Fig. \ref{fig:taylor_undamped}). We summarize our evaluations across methods and criteria in Table \ref{tab:DHOSHO_method_comparison}.

\begin{figure*}
  \includegraphics[width = 0.95\textwidth]{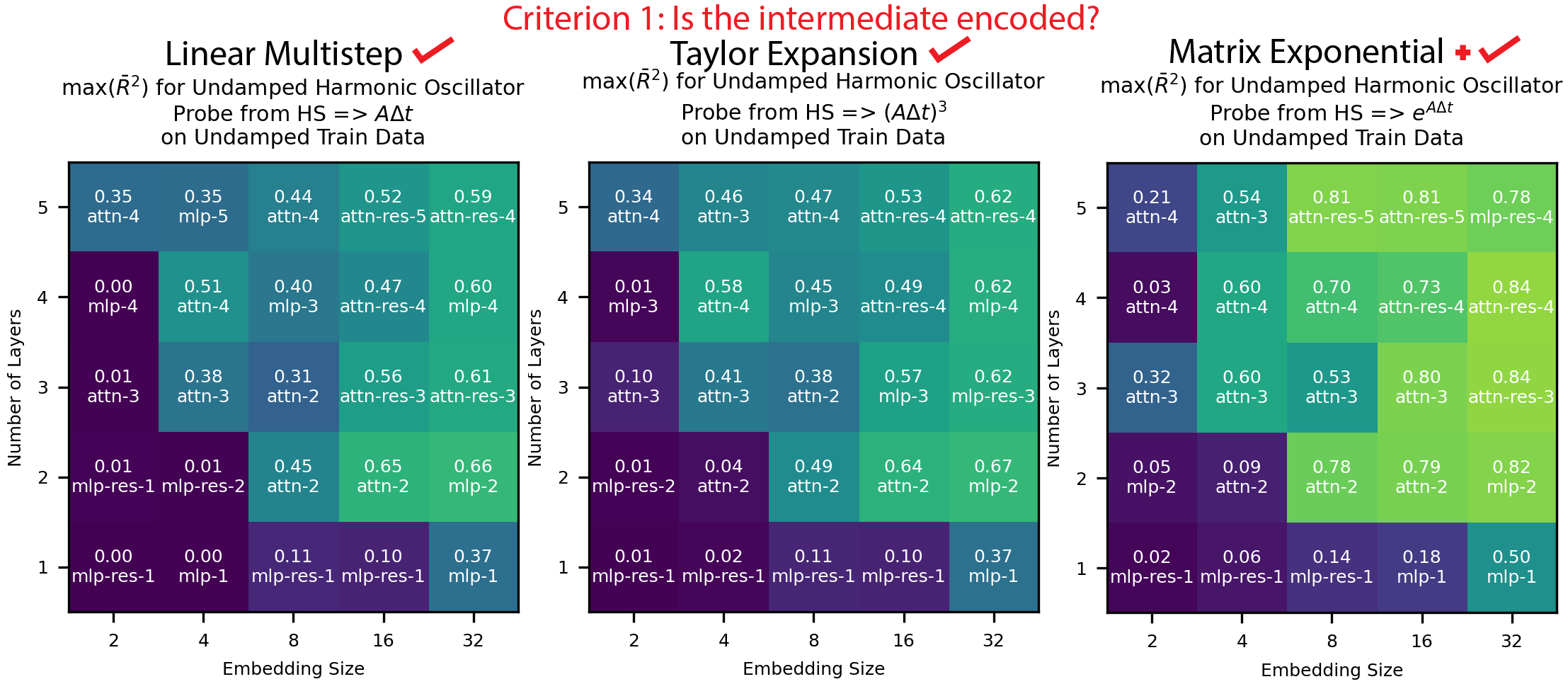}
  \caption{We analyze the intermediates of our undamped harmonic oscillator models, and find all three methods encoded, with the matrix exponential method best represented. This provides initial correlational evidence for all three methods.}
  \label{fig:undamped_crit1}
\end{figure*}

\textbf{Criterion 1: Is the intermediate encoded?} In Fig. \ref{fig:undamped_crit1}, we see that all three intermediates are well encoded in the model, with the matrix exponential method especially prominent. This provides initial correlational evidence that the models are learning numerical methods. The magnitude of encodings are generally smaller than the linear regression case, which we attribute to the increased difficulty of encoding $2 \times 2$ matrices compared to a single weight value $\bm{w}$. Notably, we only probe for linear encodings given that $\bm{w}$ was most often encoded linearly in the linear regression case study. 

\begin{figure*}
    \centering
  \includegraphics[width = 1\textwidth]{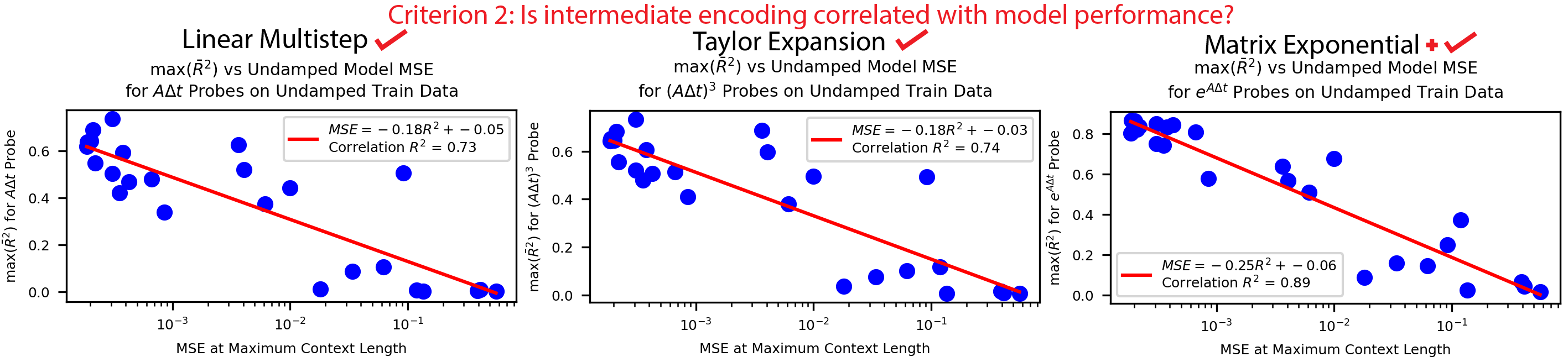}
  \caption{We find that better performing models have intermediates of all methods better encoded, but this correlation is strongest in magnitude and slope for the matrix exponential method. This is additional correlational evidence for all three methods.}
  \label{fig:undamped_crit2}
\end{figure*}

\textbf{Criterion 2: Is the intermediate encoding correlated with model performance?} In Fig. \ref{fig:undamped_crit2}, we see that for all three methods, better performing models generally have stronger encodings and worse performing models have weaker encodings. This correlation is strongest for the matrix exponential method. This provides more correlational evidence that our models are using the described methods. 

\begin{figure*}
\centering
  \includegraphics[width = 0.95\textwidth]{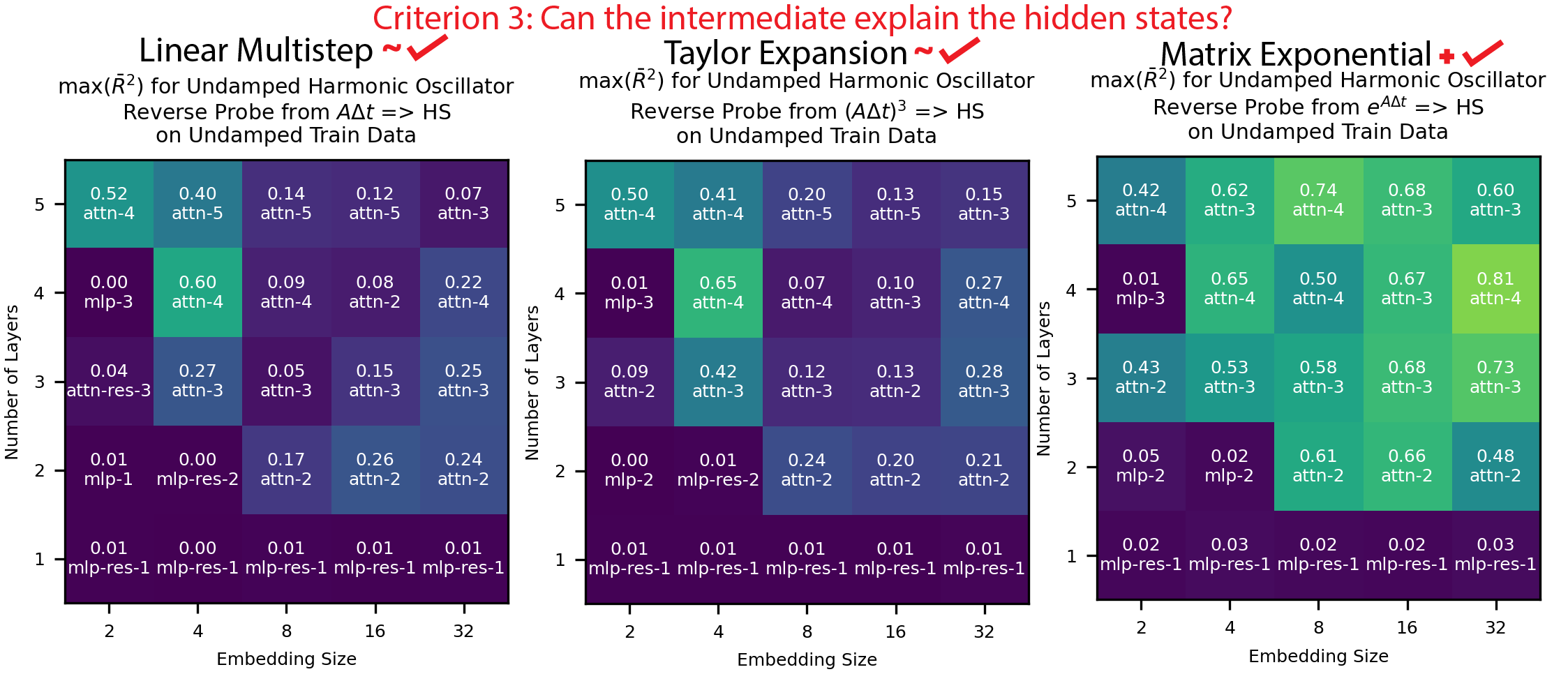}
  \caption{We find that the intermediates from all three methods can explain some variance in model hidden states, but the matrix exponential method is most consistent and successful by a wide margin.}
  \label{fig:undamped_crit3}
\end{figure*}

\textbf{Criterion 3: Can the intermediates explain the models' hidden states?} In Fig. \ref{fig:undamped_crit3}, we reverse probe from the intermediates to the models' hidden states, and find that all methods explain non trivial variance in model hidden states, while the matrix exponential method consistently explains the most variance by a sizable margin. This provides little weak causal evidence that the models are using the linear multistep and Taylor expansion methods, and stronger weak causal evidence that the models are using the matrix exponential method.

\begin{figure*}
    \centering
  \includegraphics[width = 0.95\textwidth]{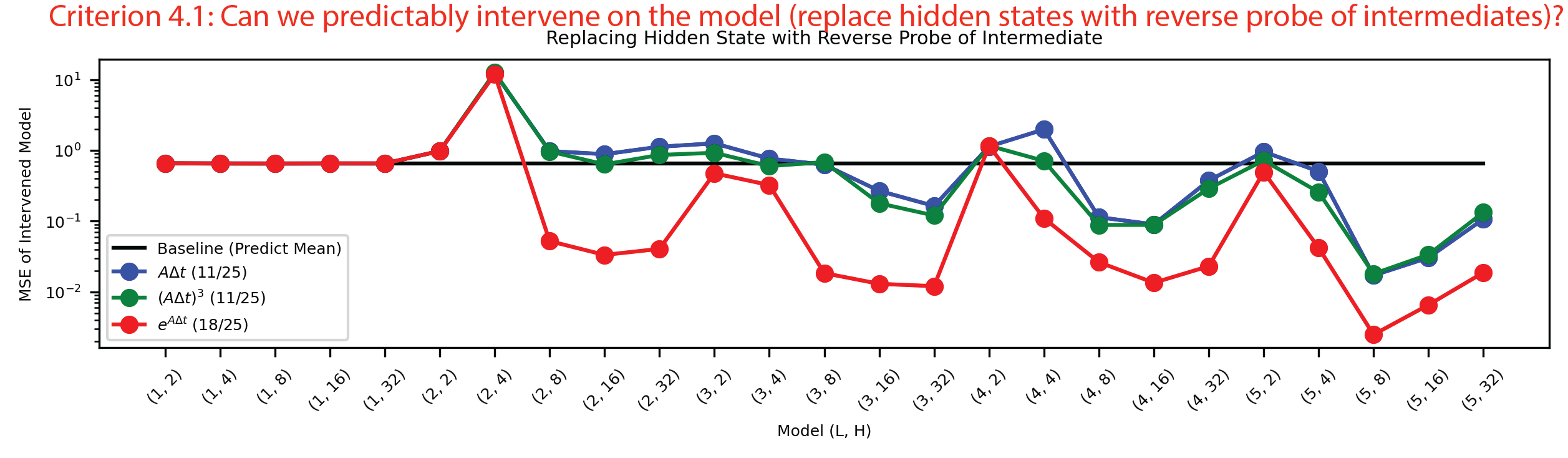}
  \caption{For each model and method, we replace the hidden state in Fig. \ref{fig:undamped_crit3} with the reverse probe of the intermediate. We see this intervention is consistently best performing for the matrix exponential method by an order of magnitude, and 18/25 models perform better than our baseline of guessing.}
  \label{fig:undamped_intervened_replace}
\end{figure*}

\textbf{Criterion 4: Can we predictably intervene on the the model?} \textit{Criterion 4.1} To intervene on the model, we use the reverse probes from Fig. \ref{fig:undamped_crit3} to generate predicted hidden states from each intermediate. In Fig. \ref{fig:undamped_intervened_replace}, we then insert these hidden states back into the model and see if the model is still able to model the SHO. The matrix exponential method has the most successful interventions by an order of magnitude, and 18/25 of these intervened models perform better than guessing. This implies that the information the transformer uses to model the SHO is stored in the matrix exponential's intermediate.

\begin{figure*}
    \centering
  \includegraphics[width = 0.85\textwidth]{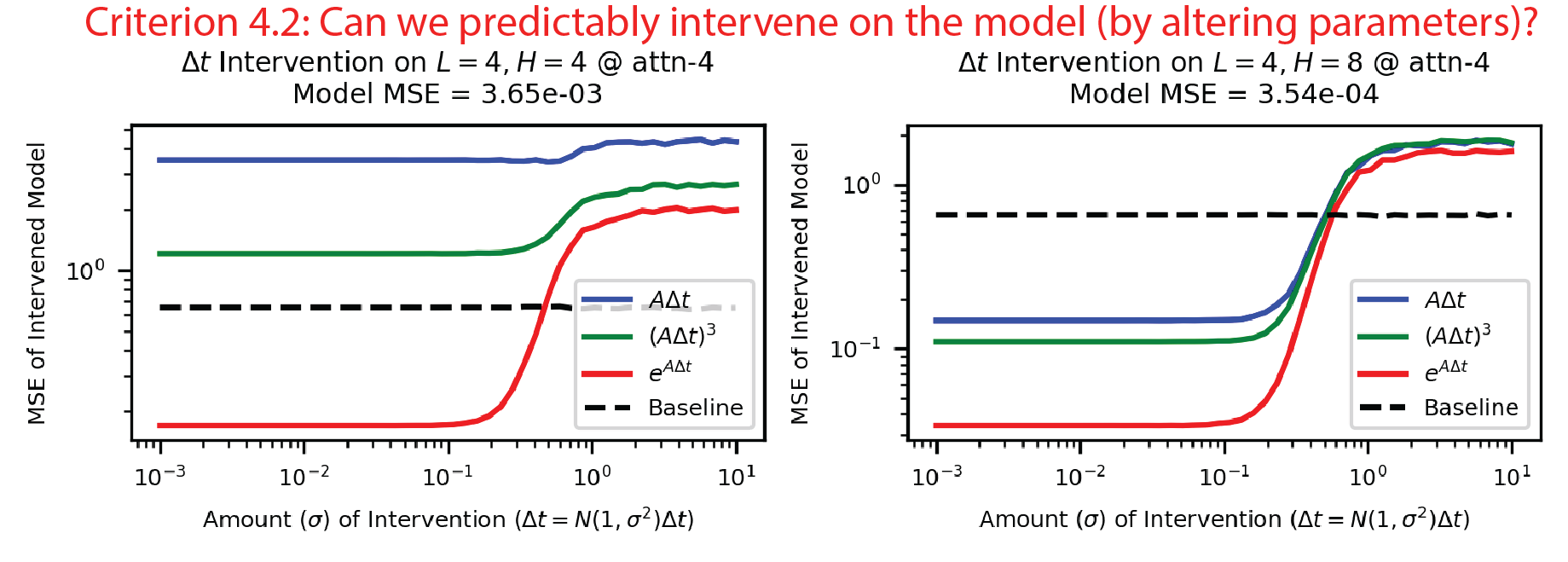}
  \caption{We vary the value of $\Delta t$ used in the intermediates and use the reverse probes from Fig. \ref{fig:undamped_crit3} to generate hidden states from these intermediates. We perform this operation on two models, which have the best linear multistep/Taylor expansion ($L = 4, H = 4$) and matrix exponential ($L = 4, H = 8$) reverse probes respectively, and find that the matrix exponential is consistently most robust to interventions. The baseline is if our model only predicted the mean of the dataset.}
  \label{fig:undamped_intervention_dt}
\end{figure*}

\textit{Criterion 4.2} We can also vary $\Delta t \rightarrow \Delta t', \omega_0 \rightarrow \omega_0'$, regenerate intermediates and then hidden states, insert these modified hidden states into the model, and see if the model makes predictions as if it "believes" the input SHO data uses $\Delta t', \omega'$. In Fig. \ref{fig:undamped_intervention_dt}, we perform this intervention on $\Delta t$, but our results are robust to intervening on $\omega_0$ as well (Appendix Fig. \ref{fig:undamped_intervention_all}). Even for the model with the best reverse probe quality for the linear multistep/Taylor expansion intermediates ($L = 4, H = 4$), intervening with the matrix exponential method is most successful. Combined with our previous intervention (4.1), we now have strong causal evidence for the matrix exponential method.

\textbf{The transformer likely uses the matrix exponential to model the undamped harmonic oscillator}
We have correlational evidence that the model is using all three methods in our theory hub, with little causal evidence for the linear multistep and Taylor expansion methods, and strong causal evidence for the matrix exponential method. We suspect the model is \textit{only} using the matrix exponential method in its computations, and the evidence we have for the other two methods is a byproduct of the use of the matrix exponential. In Appendix Fig. \ref{fig:crit13_ifexp}, we give correlational evidence for this claim by generating synthetic hidden states from $e^{\bm{A} \Delta t}$ and showing that in this synthetic setup, we retrieve values for criterions 1, 3 for linear multistep and Taylor expansion that are close to those we observe in Table \ref{tab:DHOSHO_method_comparison}.

Thus, we conclude that the transformer is likely using the matrix exponential method. This makes sense given the problem setting - both the linear multistep and Taylor expansion methods are only accurate for small $\Delta t$, while our bound of $\Delta t = U[0, 2\pi/\omega_0]$ violates this assumption for some timeseries. Still, it is remarkable that transformers use a known numerical method to model the undamped harmonic oscillator, and we can provide evidence for its use, although our experiments do not rule out the possibility of other methods being used in conjunction with the matrix exponential.

\newcommand{\bl}[1]{\textcolor{blue}{#1}}
\newcommand{\rl}[1]{\textcolor{red}{#1}}
\newcommand{\gl}[1]{\textcolor{orange}{#1}}

\begin{table}[h!]
\centering
\small
\begin{tabular}{cccc}
\textbf{Criterion} & \textbf{Linear Multistep} & \textbf{Taylor Expansion} & \textbf{Matrix Exponential} \\
1. Intermediate Encoding & $\bl{0.66}/\rl{0.51}$      & $\bl{0.67}/\rl{0.25}$      & $\bm{\bl{0.84}}/\bm{\rl{0.54}}$       \\
2. Performance, Encoding Correlation &  $\bl{0.73}/\bm{\rl{0.44}}$   & $\bl{0.74}/\rl{0.39}$ & $\bm{\bl{0.89}}/\bm{\rl{0.44}}$            \\
3. Intermediate's Explanatory Power & $\bl{0.42}/\rl{0.15}$ & $\bl{0.53}/\rl{0.11}$    & $\bm{\bl{0.78}}/\bm{\rl{0.16}}$\\
4. Intervention Success & $\bl{0.44}/\rl{X}$ & $\bl{0.44}/\rl{X}$      & $\bm{\bl{0.72}}/\rl{X}$       \\
\end{tabular}
\caption{We summarize the evaluation of methods and criteria for the \bl{undamped}/\rl{underdamped} models. For each criteria, we list a single quantity for readability. Criterion 1 is 
the largest value in Fig. \ref{fig:undamped_crit1}, criterion 2 is 
the correlation in Fig. \ref{fig:undamped_crit2}, criterion 3 is 
the largest value in Fig. \ref{fig:undamped_crit3}, and criterion 4 is the ratio in the legend of Fig. \ref{fig:undamped_intervened_replace}. The matrix exponential performs best across criteria.}
\label{tab:DHOSHO_method_comparison}
\end{table}

\subsection{Extension to the damped harmonic oscillator ($\gamma \neq 0$)}

We want to understand the generality of our finding by extending our problem space to the damped harmonic oscillator, where $\gamma \neq 0$. We leave relevant details about our procedure to Appendix \ref{sec:dampedspringres}, but in Table \ref{tab:DHOSHO_method_comparison}, we find our intermediate analysis performs much more poorly on the underdamped case than undamped. We describe possible explanations in Appendix \ref{sec:dampedspringres}, but because of this we temper our finding from the undamped harmonic oscillator with caution about its generality.

\section{Conclusions}
\label{sec:conclusions}
After developing criteria for intermediates in the toy setting of linear regression, we find that transformers use known numerical methods for modeling the simple harmonic oscillator, specifically the matrix exponential method. We leave the door open for researchers to better understand the methods transformers use to model the damped harmonic oscillator and use the study of intermediates to understand how transformers model other systems in physics.

\textbf{Limitations} We analyze relatively small transformers with only one attention head and no LayerNorm. While we demonstrate strong results for the undamped harmonic oscillator, our results for the underdamped harmonic oscillator are more mild. We only use noiseless data.



\section{Acknowledgements}
We thank Wes Gurnee, Isaac Liao, Josh Engels, and Vedang Lad for fruitful  discussions and MIT SuperCloud \cite{reuther2018interactive} for providing computation resources.  This work is supported by the Rothberg Family Fund for Cognitive Science, the NSF Graduate Research Fellowship (Grant No.
2141064), and IAIFI through NSF grant PHY-2019786.

\bibliography{ref}
\bibliographystyle{unsrt}

\clearpage

\appendix

\begin{center}
    {\LARGE\bf  Supplementary material}
\end{center}

\section{Related Work}
\label{sec:relatedwork}

\textbf{Mechanistic interpretability} Mechanistic interpretability (MI) as a field aims to understand the specific computational procedures machine learning models use to process inputs and produce outputs \cite{elhage2022superposition, olsson2022context, nanda2023progress, liu2022towards,elhage2021mathematical, chughtai2023toy, gurnee2023finding, wang2023interpretability, conmy2023towards}. Some MI work focuses on decoding the purpose of individual neurons \cite{Gurnee2023LanguageMR} while other work focuses on ensembles of neurons \cite{wang2023interpretability, conmy2023towards}. Our work is aligned with the latter.

\textbf{Algorithmic behaviors in networks} A subset of MI attempts to discover the specific algorithms networks use to solve tasks by reverse engineering weights. For example, it has been demonstrated that transformers use the discrete Fourier transform to model modular addition \cite{nanda2023progress}, while additional work has found competing hypotheses for the specific algorithm transformers use on this task\cite{zhog2023clock}. Instead of reverse engineering weights, we make use of linear probing \cite{alain2016probe} to discover byproducts of algorithms represented internally by transformers. Studies have found that algorithms in models are potentially an "emergent" behavior that manifests with size \cite{wei2022emergent, schaeffer2023emergent}, which we also find. 

\textbf{AI \& Physics} Many works design specialized machine learning architectures for physics tasks \cite{Udrescu2020feynman, liu2021poincare, cranmer2020lagrangian, greydanus2019hamiltonian, Kantamneni2024OptPDEDN}, but less work has been done to see how well transformers perform on physical data out of the box. Recently, it was shown that LLMs can in-context learn physics data \cite{liu2024physcontext}, which inspired the research question of this paper: how do transformers model physics?

\section{Additional Results for Linear Regression}
\label{sec:linregappendix}

\begin{figure}[ht]
  \centering

  \includegraphics[width = 0.7\textwidth]{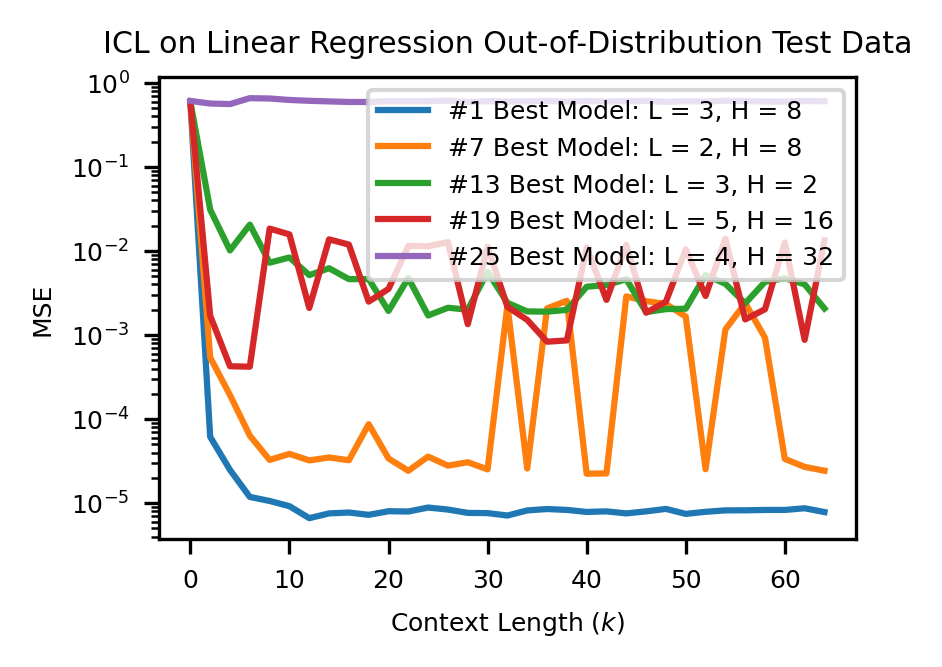}
  \caption{We find that linear regression models are able to generalize to out-of-distribution test data with $0.75\leq|w| \leq 1$.}
  \label{fig:linregtest}

\end{figure}

\begin{figure}[ht]
  \centering

  \includegraphics[width = 1\textwidth]{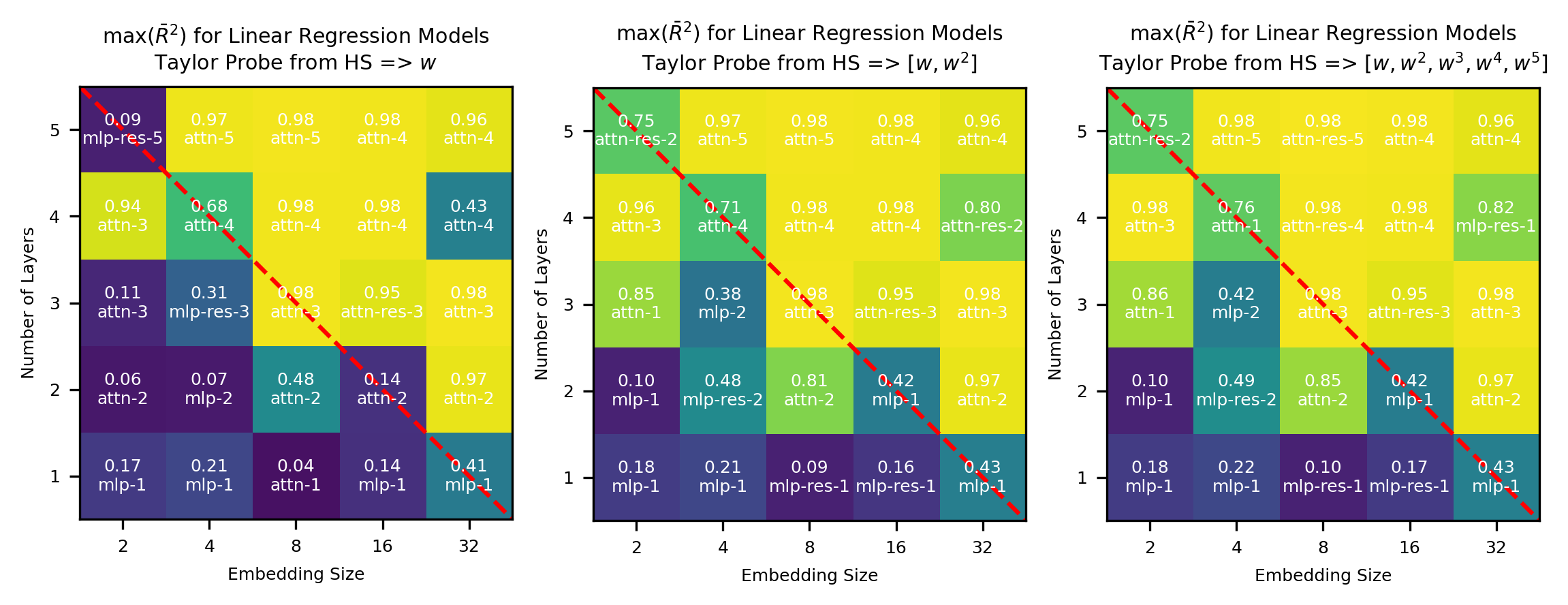}
  \caption{We calculate the mean of the $R^2$ of probes for $f(w)$ across all layers of the transformer and annotate each model with its highest mean score, $\max(\bar{R^2})$. When $f(w)$ is linear (left) and quadratic (middle), we observe a striking phase transition of encoding based on model size, demarked by the red dashed line. If $\bm{w}$ is encoded, it is mostly encoded linearly, with the $(L,H) = (5,2), (4,32), (2,8)$ models showing signs of a quadratic representation of $\bm{w}$. We do not see any meaningful gain in encoding when extending the Taylor probe to degree $n>2$ (right). For models where $f(w)$ is well represented, it often happens in an attention layer. This is possibly because the attention layer aggregates all past estimates of $f(w)$ into an updated estimate.}
  \label{fig:phasetransition}

\end{figure}

\begin{figure}[ht]
  \centering

  \includegraphics[width = 0.6\textwidth]{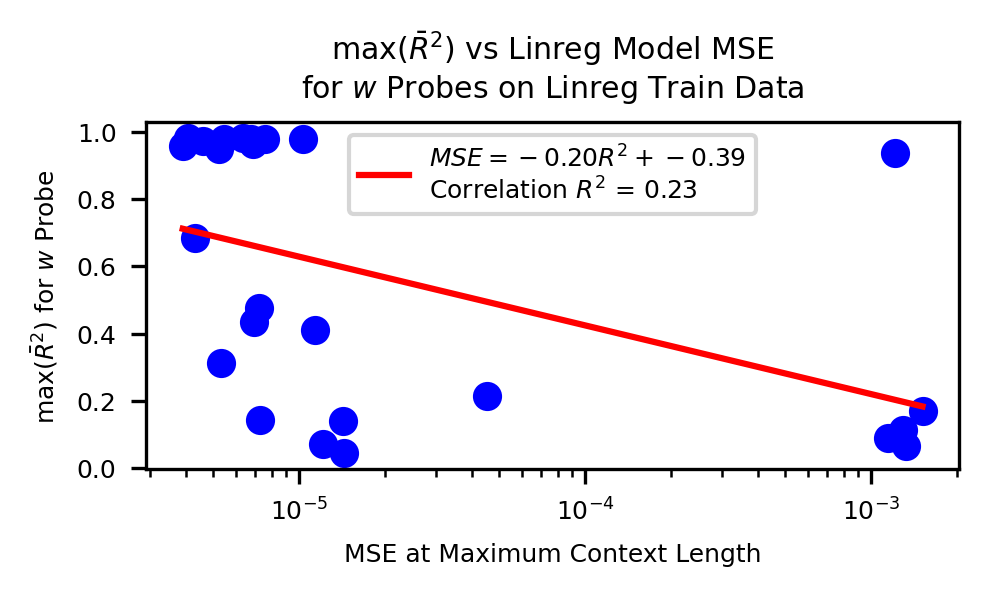}
  \caption{Better performing models generally have better encodings of $\bm{w}$, while worse performing models generally have worse encodings (other than one outlier in the top right)}
  \label{fig:linregmser2corr}

\end{figure}

In Fig. \ref{fig:linregtest}, we find that our transformers are able to generalize to linear regression test examples with out-of-distribution data ($0.75\leq|w| \leq 1$). In Fig. \ref{fig:phasetransition} we see that smaller models do not have $\bm{w}$ encoded, while larger models often have $\bm{w}$ linearly encoded (with some quadratic encodings as well). In Fig. \ref{fig:linregmser2corr}, we see that better performing models generally have better encodings, while worse performing models generally have worse encodings.

\section{Theory hub generalizes to other systems}
\label{sec:generalizingtoothersystems}
We note that the theory hub we summarize in Table \ref{tab:num_methods} is valid for all differential equations that can be written as $\dot{\bm{x}} = \bm{Ax}$ if $\bm{A}$ is a constant matrix. This includes all homogenous linear differential equations with constant coefficients, and potentially non linear differential equations as well. Koopman operator theory allows nonlinear differential equations to be modeled as linear differential equations. Here is an example taken from \cite{Brunton2019NotesOK}:

\begin{mdframed}[innertopmargin=10pt, innerbottommargin=10pt, innerleftmargin=10pt, innerrightmargin=10pt, backgroundcolor=gray!10, roundcorner=10pt]
Here, we consider an example system with a single fixed point, given by:
\begin{equation}
    \dot{x}_1 = \mu x_1 \tag{32a}
\end{equation}
\begin{equation}
    \dot{x}_2 = \lambda (x_2 - x_1^2). \tag{32b}
\end{equation}
For $\lambda < \mu < 0$, the system exhibits a slow attracting manifold given by $x_2 = x_1^2$. It is possible to augment the state $x$ with the nonlinear measurement $g = x_1^2$, to define a three-dimensional Koopman invariant subspace. In these coordinates, the dynamics become linear:
\begin{equation}
    \frac{d}{dt} \begin{bmatrix} y_1 \\ y_2 \\ y_3 \end{bmatrix} = \begin{bmatrix} \mu & 0 & 0 \\ 0 & \lambda & -\lambda \\ 0 & 0 & 2\mu \end{bmatrix} \begin{bmatrix} y_1 \\ y_2 \\ y_3 \end{bmatrix} \quad \text{for} \quad \begin{bmatrix} y_1 \\ y_2 \\ y_3 \end{bmatrix} = \begin{bmatrix} x_1 \\ x_2 \\ x_1^2 \end{bmatrix}. \tag{33a}
\end{equation}
\end{mdframed}
For this nonlinear system, our theory hub in Table \ref{tab:num_methods} is still relevant using 
$$\bm{A} = \begin{bmatrix} \mu & 0 & 0 \\ 0 & \lambda & -\lambda \\ 0 & 0 & 2\mu \end{bmatrix}, \bm{x} = \begin{bmatrix} x_1 \\ x_2 \\ x_1^2 \end{bmatrix}.$$ 
Thus, it is possible that the methods we've determined a transformer uses to model the simple harmonic oscillator extends to other, more complex systems.

\section{Undamped Harmonic Oscillator Appendices}
\label{sec:uho_app}

\textbf{Data generation for the undamped harmonic oscillator} We generate 5000 sequences of 65 timesteps for various values of $\omega_0, \Delta t, x_0, $ and $v_0$. We range $\omega_0  = U[\frac{\pi}{4}, \frac{5\pi}{4}]$, $\Delta t = U[0, \frac{2\pi}{ \omega_0}]$, $x_0, v_0 = U[-1,1]$. The undamped harmonic oscillator is periodic so using a larger $\Delta t$ is not useful. We also generate an out-of-distribution test set with $\omega_0 = U[0,\frac{\pi}{4}]+U[\frac{5\pi}{4},\frac{3\pi}{2}]$ with the same size as the training set.

\begin{figure*}
    \includegraphics[width = 1\textwidth]{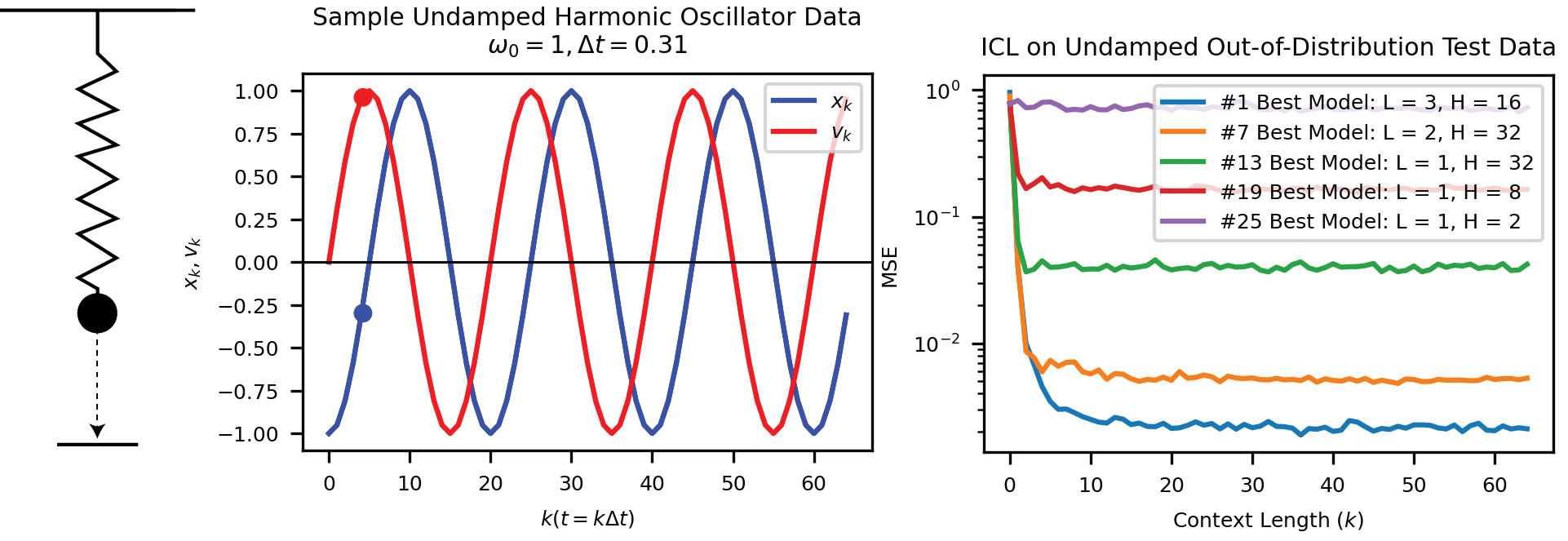}
    \caption{An intuitive picture for a simple harmonic oscillator is a mass oscillating on a spring (left). The trajectory of the SHO can be fully parameterized by the value of $x,v$ at various timesteps (middle), and we find that models trained to predict undamped SHO trajectories are able to generalize to out-of-distribution test data with in-context examples (right).}
    \label{fig:undampedSHOdata}
\end{figure*}

\begin{figure*}
    \centering
    \includegraphics[width = 1\textwidth]{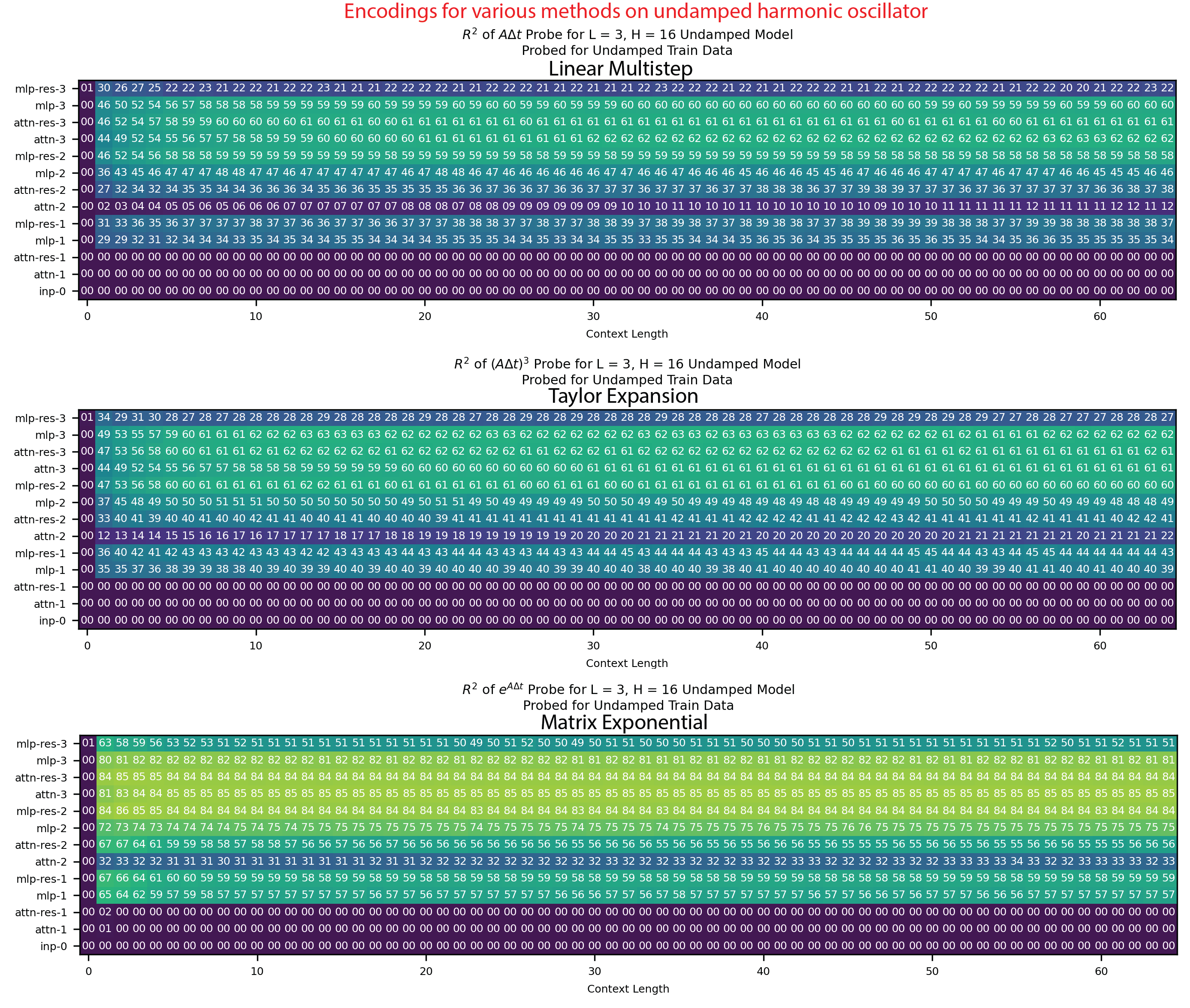}
    \caption{We visualize the evolution of encodings across all methods with context length for the best performing undamped model.}
    \label{fig:undamped_intermediate_decay}
    
\end{figure*}

\begin{figure*}
    \centering
    \includegraphics[width = 1\textwidth]{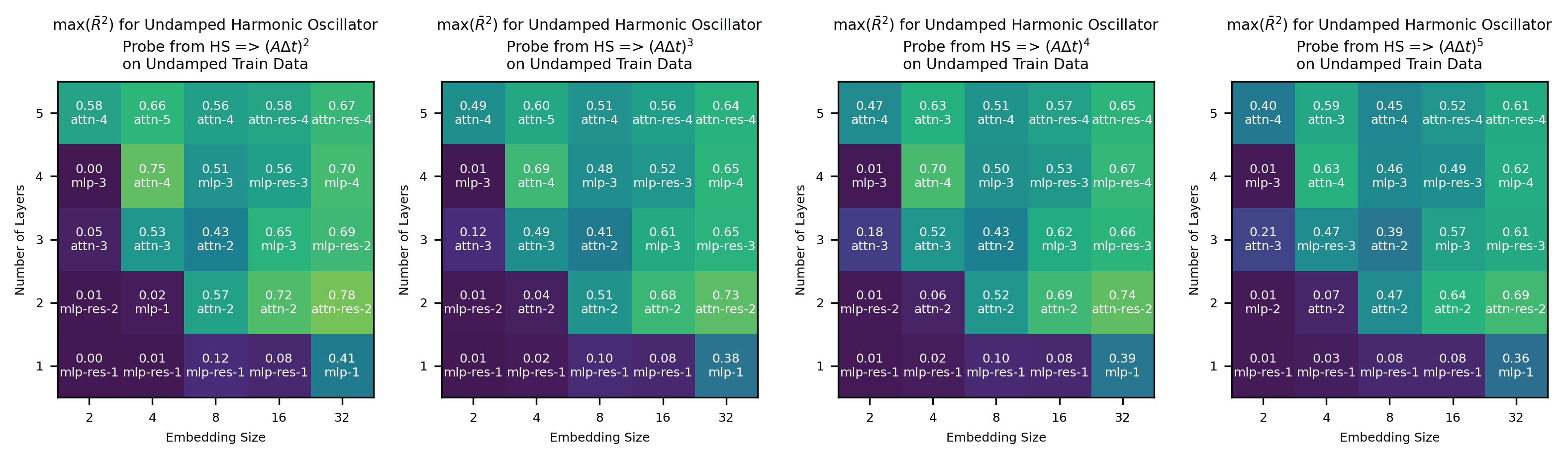}
    \caption{We find that our choice of $j$ in the intermediate for the Taylor expansion method ($(A \Delta t)^j$ has little effect on our results or conclusions about the undamped harmonic oscillator (shown for criterion 1).}
    \label{fig:taylor_undamped}
    
\end{figure*}

\begin{figure*}
    \centering
    \includegraphics[width = 1\textwidth]{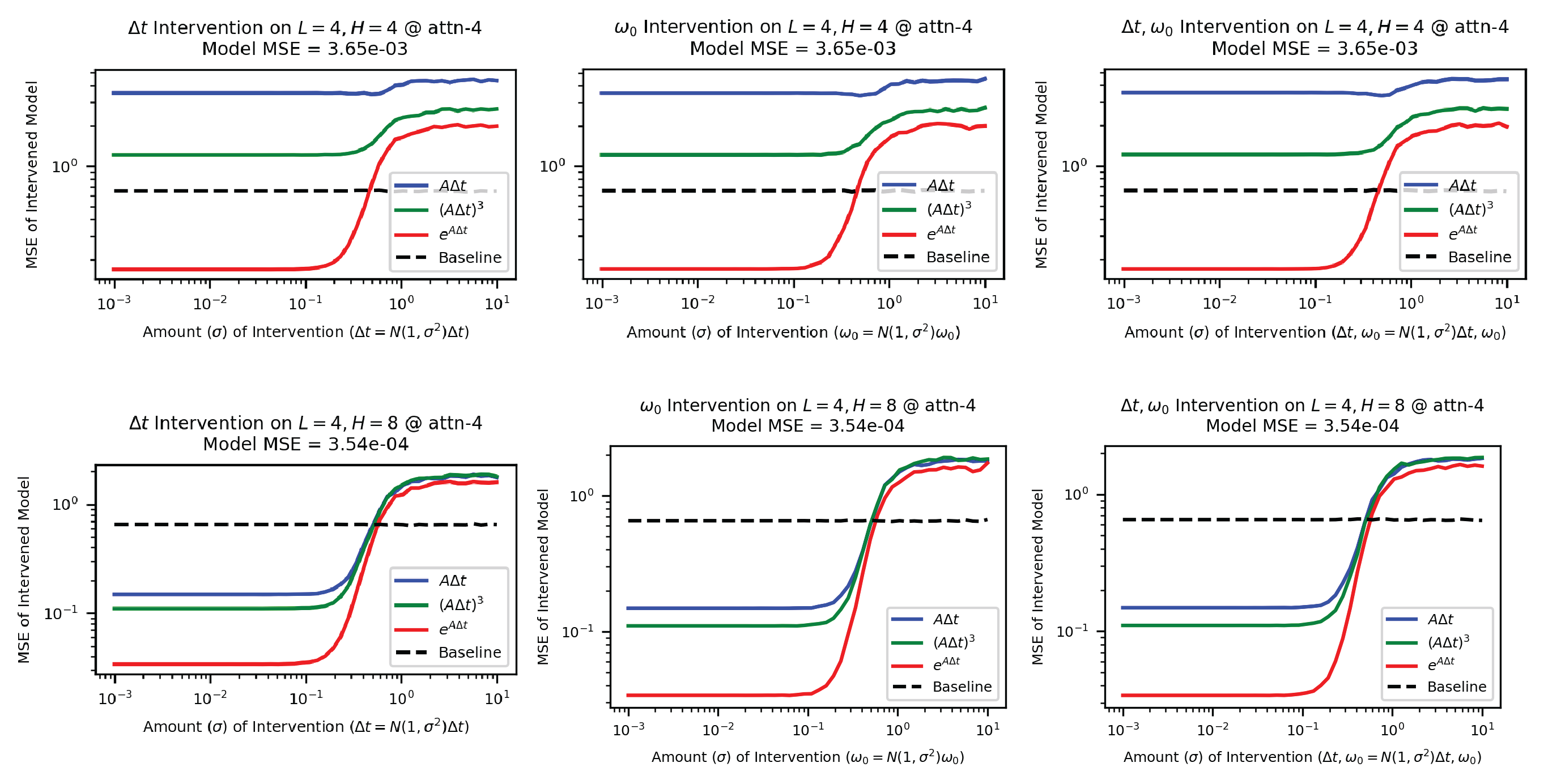}
    \caption{Regardless of which quantities we intervene on, our general results are robust for criterion 4 for the undamped harmonic oscillator.}
    \label{fig:undamped_intervention_all}
    
\end{figure*}

\begin{figure*}
    \centering
    \includegraphics[width = 1\textwidth]{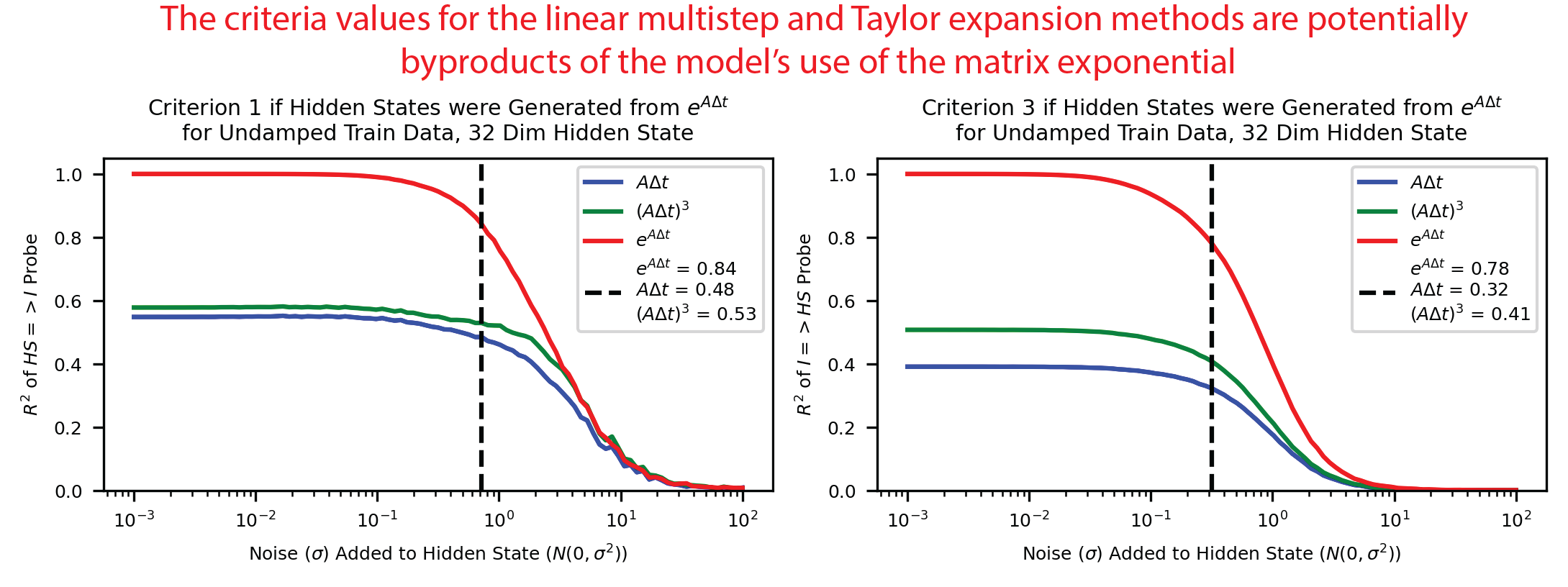}
    \caption{We generate synthetic hidden states from the matrix exponential intermediates and find that this naturally results in values for criterion 1,3 for the linear multistep and Taylor expansion methods that are close to those we observe in Table \ref{tab:DHOSHO_method_comparison}. This is 
 correlational evidence that the matrix exponential method is potentially solely used by the transformer, and values for the other two methods are byproducts. These byproducts could arise because $e^{A \Delta t} = \sum_j (A \Delta t)^j/j!$}
    \label{fig:crit13_ifexp}
    
\end{figure*}

\textbf{Additional results for undamped harmonic oscillator} In Fig. \ref{fig:undampedSHOdata}, we see that models are able to learn the undamped harmonic oscillator in-context, even for values of $\omega_0$ out of the distribution these models were trained on. We also plot the evolution of encodings for our various methods on the best performing undamped model in Fig. \ref{fig:undamped_intermediate_decay}. We find that our choice of $j$ for the Taylor expansion method has is mostly irrelevant for $j \leq 5$ in Fig. \ref{fig:taylor_undamped}. With respect to criterion 4, in Fig. \ref{fig:undamped_intervention_all} we show that our intervention results are robust to which parameter we're intervening on ($\Delta t, 
\omega_0$, or both). We also generate synthetic hidden states from the matrix exponential intermediate and find that the values for criterion 1,3 for the other two methods are potentially byproducts of the matrix exponential in Fig. \ref{fig:crit13_ifexp}, giving additional correlational evidence that the matrix exponential is the dominant method of the transformer. 

\section{Investigating the damped harmonic oscillator ($\gamma > 0$)}

\label{sec:dampedspringres}

\subsection{Mathematical setup}
The damped harmonic oscillator has three well studied modes: underdamped, overdamped, and critically damped cases. The underdamped case occurs when $\gamma < \omega_0$, and represents a spring oscillating before coming to rest. The overdamped case occurs when $\gamma > \omega_0$, and represents a spring immediately returning to equilibrium without oscillating. The analytical equations for both cases are

\noindent 
\begin{minipage}{0.5\textwidth}
\centering
\textbf{Underdamped ($\gamma < \omega_0$)}
\small 
\begin{align*}
    \bm{x_k} &= e^{-k \gamma \Delta t} \bigg( x_0 \cos (k \omega \Delta t) + \\
         &\quad \frac{v_0+\gamma x_0}{\omega} \sin (k \omega \Delta t)\bigg) \\ 
    \bm{v_k} &= e^{-k \gamma \Delta t} \bigg(v_0 \cos (k \omega \Delta t) - \\
         &\quad \left(\frac{v_0+\gamma x_0}{\omega}\gamma + \omega x_0\right) \sin (k \omega \Delta t)\bigg)
\end{align*}
\end{minipage}%
\begin{minipage}{0.5\textwidth}
\centering
\textbf{Overdamped ($\gamma > \omega_0$)}
\small 
\begin{align*}
    \bm{x_k} &= \frac{e^{-k \gamma \Delta t}}{2} \bigg((x_0+\frac{v_0+\gamma x_0}{\omega})e^{k \omega \Delta t} + \\
         &\quad (x_0-\frac{v_0 + \gamma x_0}{\omega})e^{-k\omega \Delta t}\bigg) \\
    \bm{v_k} &= \frac{e^{-k \gamma \Delta t}}{2} \bigg((\omega - \gamma)(x_0+\frac{v_0+\gamma x_0}{\omega})e^{k \omega \Delta t} - \\
         &\quad (\omega + \gamma)(x_0-\frac{v_0 + \gamma x_0}{\omega})e^{-k \omega \Delta t}\bigg)
\end{align*}
\label{eq:DHOsols}
\end{minipage}

where $\omega = \sqrt{|\gamma^2-\omega_0^2|}$. Note that the critically damped case ($\gamma = \omega_0$) is equivalent to $\lim_{\gamma\to\omega_0^-}$ of the underdamped case and $\lim_{\gamma\to\omega_0^+}$ of the overdamped case. Thus, we focus our study on the underdamped and overdamped cases, and visualize sample trajectories of both in Appendix Fig. \ref{fig:dampedtraj}.

\subsection{Computational setup for the damped harmonic oscillator}
We use an analogous training setup to the undamped harmonic oscillator. We generate 5000 sequences of 32 timesteps for various values of $\omega_0, \gamma, \Delta t, x_0, $ and $v_0$ for both the underdamped and overdamped cases. For the underdamped and overdamped case, we range $\omega_0  = U[0.25\pi, 1.25\pi]$ and $\Delta t = U[0, \frac{2\pi}{13 \omega_0}]$. We use this sequence length and bound on $\Delta t$ to account for the periodic nature of the damped harmonic oscillator and also to ensure that the system does not decay to 0 too fast. For the underdamped case, we take $\gamma = U[0,\omega_0]$, and for the overdamped case, $\gamma = U[\omega_0, 1.5\pi]$. We also generate an out-of-distribution test set following a similar process but using $\omega_0 = U[0,0.25\pi]+U[1.25\pi,1.5\pi]$. 

\begin{figure*}
  \includegraphics[width = 1\textwidth]{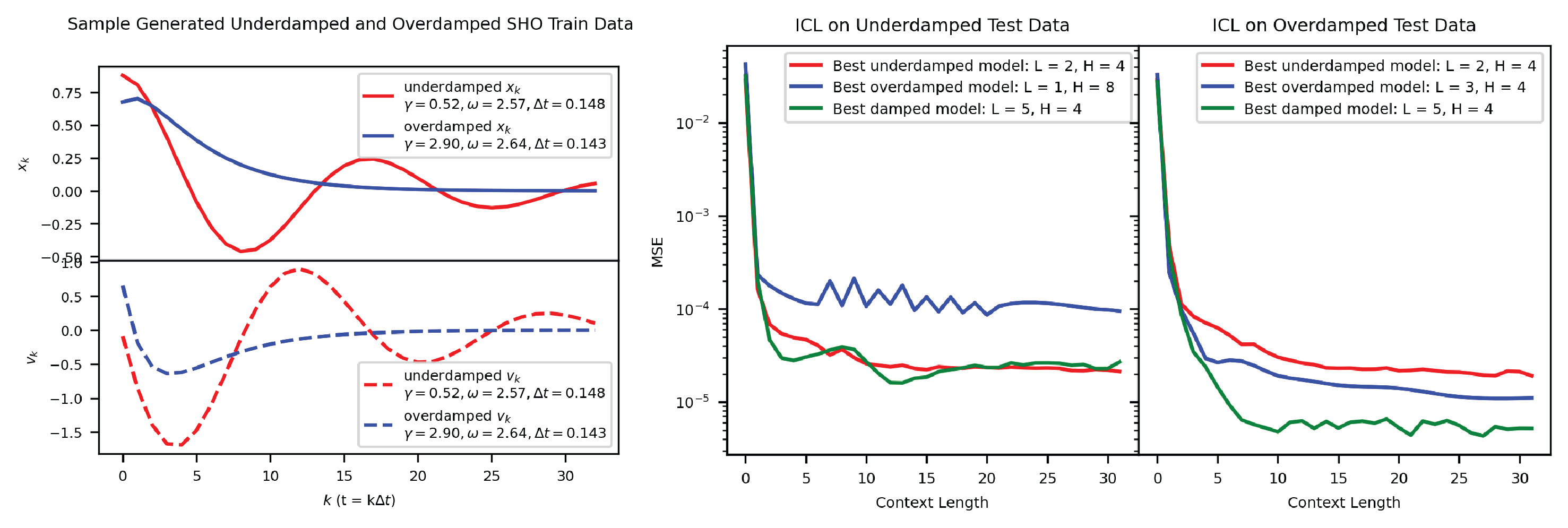}
  \caption{We generate data for underdamped and overdamped harmonic oscillators following the procedure detailed in Section 3, and visualize sample curves in the left most plot. From both the analytical equations and the plotted curves, we see that underdamped and overdamped data follow very different trajectories. Amazingly, on the right most plot we find that transformers trained on underdamped data generalize to overdamped data! This implies that our transformer is using a similar method to calculate both, otherwise this generalization would be impossible. We hypothesize that our "AI Physicist" is using one of the numerical methods from the undamped case. Note, that the "damped" oscillator was trained on equal parts underdamped and overdamped data.}
  \label{fig:dampedtraj}
\end{figure*}

In Fig. \ref{fig:dampedtraj} we find that a transformer trained on \textit{underdamped} data is able to generalize to \textit{overdamped} data with only in-context examples! This is a surprising discovery, since a human physicist who is only exposed to underdamped data would model it with the analytical function in Section \ref{eq:DHOsols}. But this method would not generalize: the underdamped case uses exponential and trigonometric functions of $\gamma \Delta t$ and $\omega \Delta t$ respectively, while the overdamped case consists solely of exponential functions. We predict that our ``AI Physicist'' is able to generalize between underdamped and overdamped cases because it is using numerical methods that model the underlying dynamics shared by both scenarios.

\subsection{Criteria are less aligned for the underdamped harmonic oscillator}
\begin{figure*}
    \includegraphics[width = 1\textwidth]{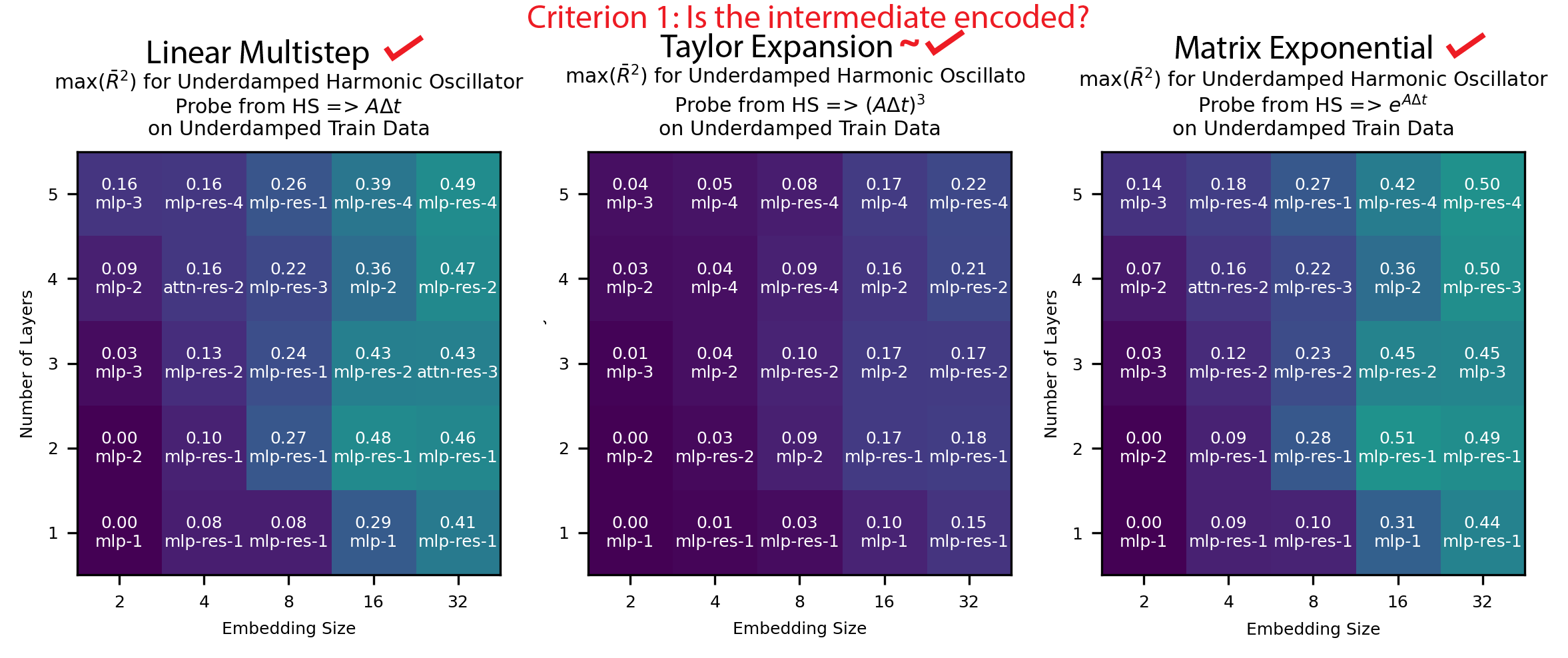}
    \caption{We observe that the intermediates for all three methods are encoded, but less than the undamped case in Fig. \ref{fig:undamped_crit1}. The linear multistep is roughly as prominent as the matrix exponential method, which is also a departure from the undamped case.}
    \label{fig:underdamped_crit1}
\end{figure*}

\begin{figure*}
    \includegraphics[width = 1\textwidth]{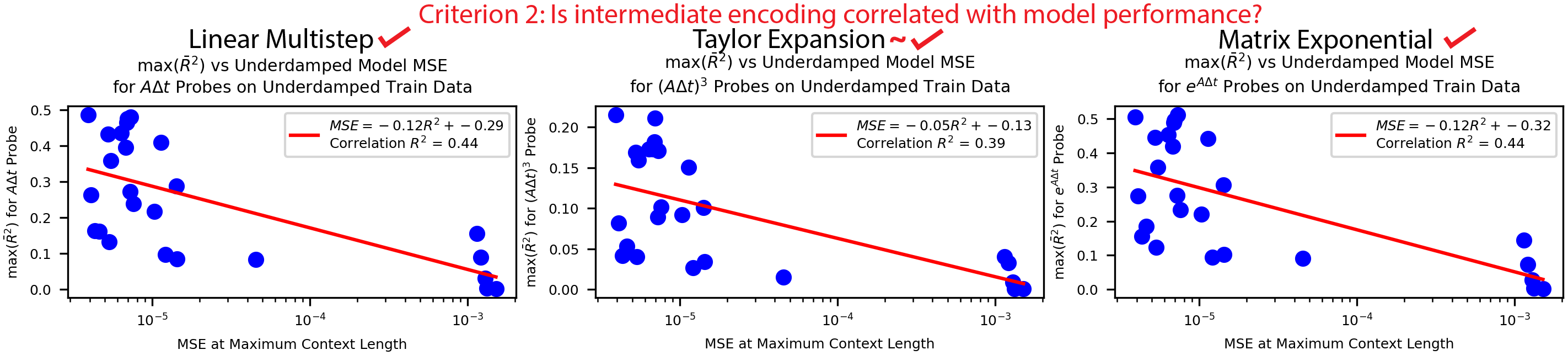}
    \caption{We see that generally, better performing models exhibit stronger encodings of intermediates, while worse performing models exhibit weaker encodings. These trends are not as strong as the undamped case, shown in Fig. \ref{fig:undamped_crit2}. Like criterion 1 in Fig. \ref{fig:underdamped_crit1}, we see that the linear multistep method is competitive with the matrix exponential method.}
    \label{fig:underdamped_crit2}
    
\end{figure*}

\begin{figure*}
    \includegraphics[width = 1\textwidth]{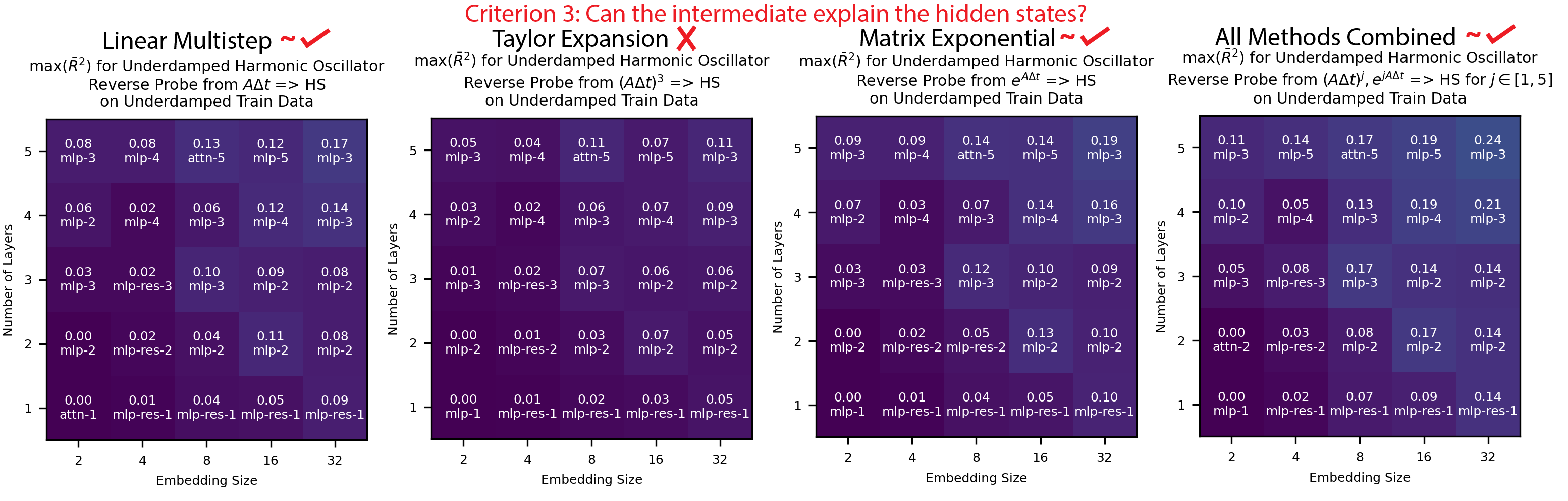}
    \caption{Multiple methods represent nontrivial amounts of variance in the hidden states, but even all methods combined (right) explain less than a quarter of the variance in the hidden states.}
    \label{fig:underdamped_crit3}
    
\end{figure*}

We evaluate all methods for the underdamped harmonic oscillator on our criteria and summarize the evaluations in Table \ref{tab:DHOSHO_method_comparison} and show relevant figures for criteria 1, 2, and 3 in Figures \ref{fig:underdamped_crit1}, \ref{fig:underdamped_crit2}, and \ref{fig:underdamped_crit3} respectively. While we see moderate correlational and some causal evidence for our proposed methods, we note that there is a steep dropoff across criteria between the undamped and underdamped cases. We identify a few possible explanations for this discrepancy:

\textbf{The transformer is using a method outside of the hypothesis space} Because the intermediates explain so little of the hidden states even when combined (Fig. \ref{fig:underdamped_crit3}), we hypothesize that the transformer has discovered a novel numerical method or is using another known method outside of our proposed hypothesis space. This is more likely for the damped case because we decrease the range on $\Delta t$ to avoid decay, which makes approximate numerical solutions more accurate. But why would it be doing this for the damped case and not the undamped case? For our damped experiments, we decrease the range on $\Delta t$ so that the trajectory does not decay to 0 to quickly, but this also allows for approximate numerical methods to be more accurate, as demonstrated by the competitive performance of the linear multistep method with the matrix exponential method in Table \ref{tab:DHOSHO_method_comparison}. So it is possible our transformer is relying on another numerical method outside of our hypothesis space.

\textbf{Natural decay requires less "understanding" by the transformer} As the context length increases, damping forces the system to naturally decay to 0, so the transformer can use less precise methods to predict the next timestep. In Appendix Fig. \ref{fig:underdamped_intermediate_decay}, we see that the intermediates' encodings accordingly decay with context length, which possibly explains the underdamped case's diminished metrics.

\textbf{More data for the transformer to encode}
With a nonzero damping factor $\gamma$, the intermediates we investigate in Table \ref{tab:num_methods} have more non-constant values in their $2 \times 2$ matrices in the damped/undamped case: the linear multistep method has 3/2 values, the Taylor expansion method has 4/2 values, and the matrix exponential method has 4/3 unique values. The increased number of non-constant values could potentially make it more difficult to properly encode intermediates.

\begin{figure*}
    \centering
    \includegraphics[height = 0.9\textheight]{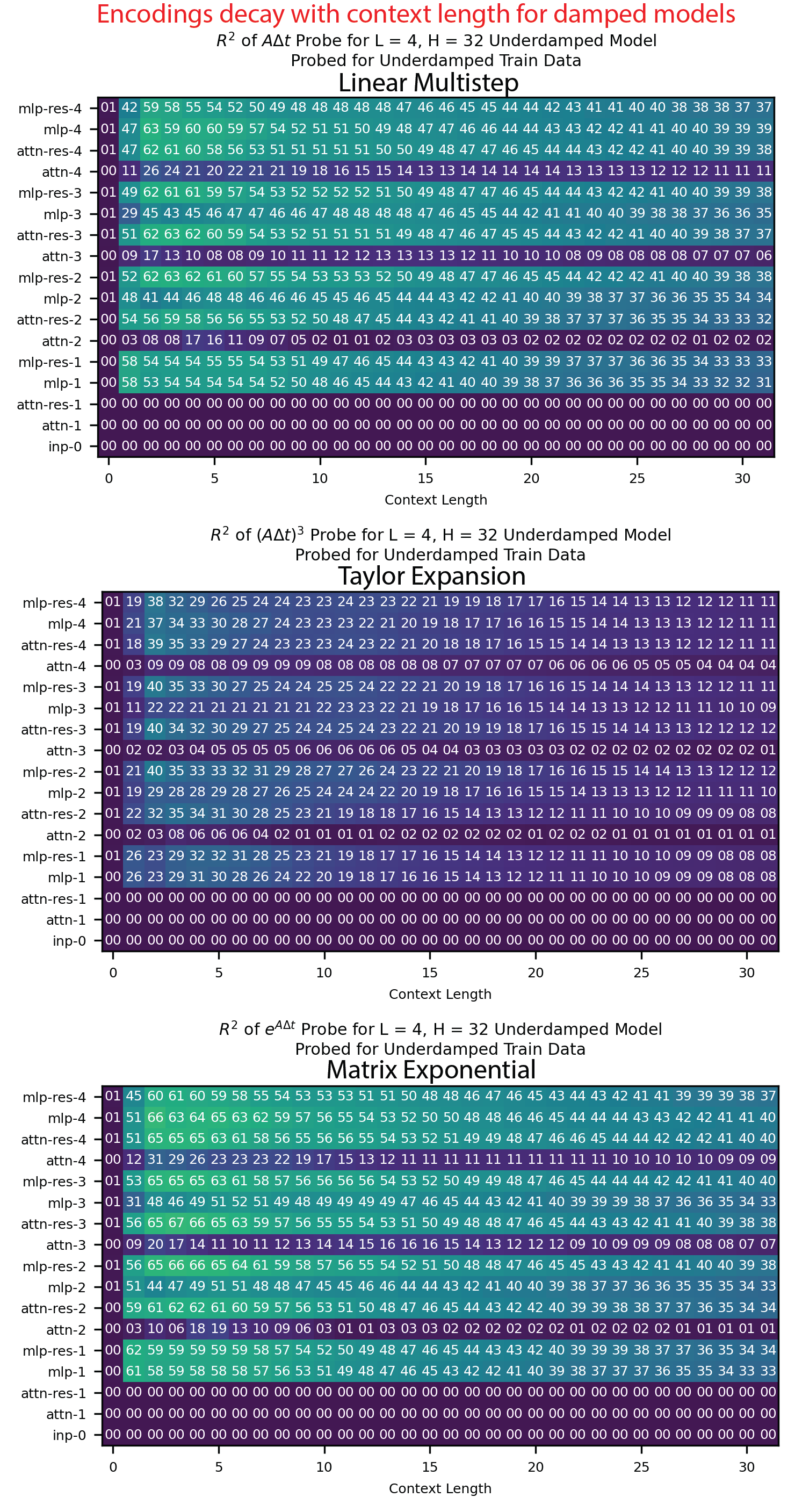}
    \caption{We see that the encoding strength of intermediates decays across all methods with context length. This similarly matches the natural decay to 0 of the damped harmonic oscillator, and is one potential explanation for why our methods are not as prominent in the damped vs undamped cases, for which the encoding quality does not decay with context length (Fig. \ref{fig:underdamped_intermediate_decay}). While this is a general observation across models, we visualize the $L = 4$, $H = 32$ model because it has the strongest encoding of intermediates from Fig. \ref{fig:underdamped_crit1}.}
    \label{fig:underdamped_intermediate_decay}
    
\end{figure*}

\begin{figure*}
    \centering
    \includegraphics[width = 1\textwidth]{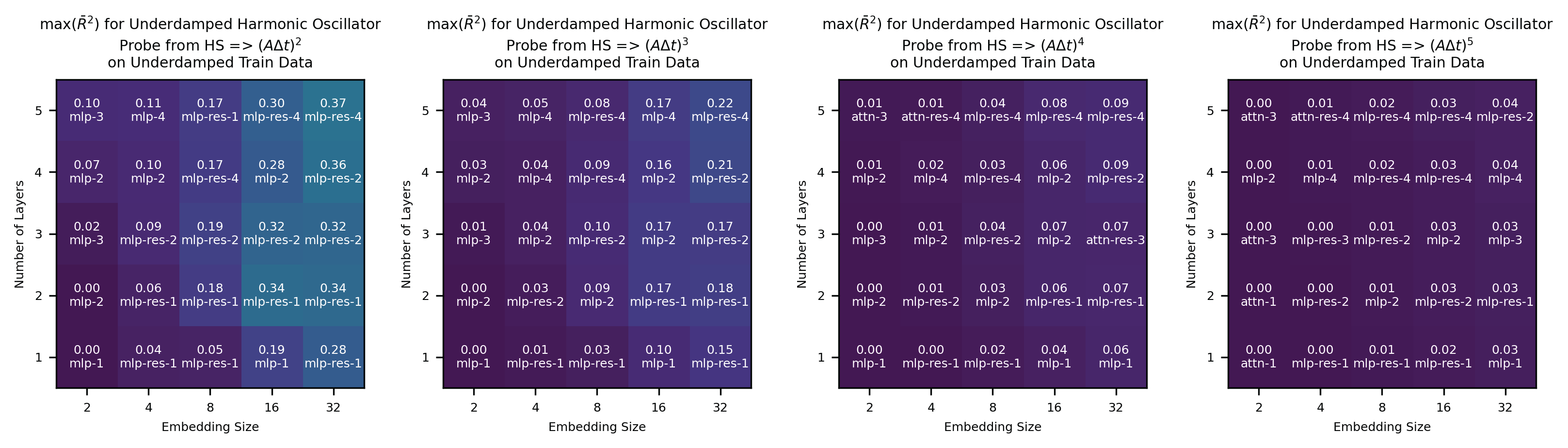}
    \caption{We find that our choice of $j$ in the intermediate for the Taylor expansion method ($(A \Delta t)^j$ has a major effect on the encoding quality, unlike the undamped case visualized in Fig. \ref{fig:taylor_undamped}. We see $j>3$ is very poorly represented in the transformer, which implies that if the transformer was using the Taylor expansion for the underdamped spring, it would likely be of order $k=3$ or less.}
    \label{fig:taylor_underdamped}
    
\end{figure*}

We leave the problem of understanding the damped harmonic oscillator to future work with intermediates.




\medskip

\end{document}